\documentclass[twocolumn,pre,floatfix,superscriptaddress,10pt, aps]{revtex4-2}
\pdfoutput=1

\usepackage[T1]{fontenc}
\usepackage[latin9]{inputenc}
\usepackage{mathrsfs}
\usepackage{amsmath}
\usepackage{amssymb}
\usepackage{graphicx}
\usepackage{graphics}
\usepackage{booktabs}
\usepackage{braket}
\usepackage{color}
\usepackage[svgnames]{xcolor}
\usepackage{comment}
\usepackage{natbib}
\usepackage{dsfont}
\usepackage{bm}
\usepackage{enumitem}
\usepackage{nicefrac}
\usepackage{pgf}
\usepackage{lmodern}
\usepackage{neuralnetwork}
\usepackage{tabularx}

\usepackage{ragged2e} 

\usepackage{subcaption}
\DeclareCaptionJustification{justified}{\justifying}
\DeclareMathOperator*{\argmin}{argmin}

\usepackage{mathtools}

\makeatletter

\usepackage{tikz}
\usetikzlibrary{arrows.meta}

\newcommand{\codevar}[1]{\texttt{#1}}

\newcommand{\deltaw}{\bm{\delta w}}

\definecolor{g1}{HTML}{D3D3D3}
\definecolor{g2}{HTML}{EEEEEE}

\usepackage{listings}
\usepackage{hyperref}
\usepackage[capitalise]{cleveref} 

\hypersetup{
  colorlinks=true,
  linkcolor=blue,
  citecolor=ForestGreen,
  urlcolor=DarkOrchid
}

\newlist{questions}{enumerate}{2}
\setlist[questions,1]{label=\textbf{RQ\arabic*.},ref=RQ\arabic*}
\setlist[questions,2]{label=(\alph*),ref=\thequestionsi(\alph*)}

\crefname{listing}{Algorithm}{Algorithms}
\Crefname{listing}{Algorithm}{Algorithms}

\newtheorem{definition}{Definition}

\AtBeginDocument{
  
}

\makeatother

\usepackage{babel}

\begin{document}

\title{Scalable iterative pruning of large language and vision models \\ using block coordinate descent}

\author{Gili~Rosenberg}
\affiliation{Amazon Quantum Solutions Lab, Seattle, WA 98170, USA}
\thanks{Corresponding author: gilir@amazon.com\\ \\ Fidelity Public Information}

\author{J.~Kyle~Brubaker}
\affiliation{Amazon Quantum Solutions Lab, Seattle, WA 98170, USA}

\author{Martin~J.~A.~Schuetz}
\affiliation{Amazon Quantum Solutions Lab, Seattle, WA 98170, USA}
\affiliation{AWS Center for Quantum Computing, Pasadena, CA 91125, USA}

\author{Elton~Yechao~Zhu}
\affiliation{Fidelity Center for Applied Technology,
FMR LLC, Boston, MA 02210, USA}

\author{Serdar~Kad{\i}o\u{g}lu}
\affiliation{AI Center of Excellence, FMR LLC, Boston, MA 02210, USA}

\author{Sima~E.~Borujeni}
\affiliation{Fidelity Center for Applied Technology,
FMR LLC, Boston, MA 02210, USA}

\author{Helmut~G.~Katzgraber}
\affiliation{Amazon Quantum Solutions Lab, Seattle, WA 98170, USA}

\date{\today}

\begin{abstract}
Pruning neural networks, which involves removing a fraction of their weights, can often maintain high accuracy while significantly reducing model complexity, at least up to a certain limit. We present a neural network pruning technique that builds upon the Combinatorial Brain Surgeon, but solves an optimization problem over a subset of the network weights in an iterative, block-wise manner using block coordinate descent. The iterative, block-based nature of this pruning technique, which we dub ``iterative Combinatorial Brain Surgeon'' (iCBS) allows for scalability to very large models, including large language models (LLMs), that may not be feasible with a one-shot combinatorial optimization approach. When applied to large models like Mistral and DeiT, iCBS achieves higher performance metrics at the same density levels compared to existing pruning methods such as Wanda. This demonstrates the effectiveness of this iterative, block-wise pruning method in compressing and optimizing the performance of large deep learning models, even while optimizing over only a small fraction of the weights. Moreover, our approach allows for a quality-time (or cost) tradeoff that is not available when using a one-shot pruning technique alone. The block-wise formulation of the optimization problem enables the use of hardware accelerators, potentially offsetting the increased computational costs compared to one-shot pruning methods like Wanda. In particular, the optimization problem solved for each block is quantum-amenable in that it could, in principle, be solved by a quantum computer. 
\end{abstract}

\date{\today}

\maketitle

\section{Introduction}
\label{sec:introduction}

Many state-of-the-art machine learning (ML) models, particularly in Deep Learning (DL), have an enormous number of parameters. In particular, many large language models (LLMs) have billions of parameters, with some even exceeding a trillion parameters \cite{fedus2022switch}. The surge in the size of ML models is partially due to the observation that their performance tends to improve with size \cite{kaplan2020scaling}, as well as the easy and (relatively) cheap availability of large compute clusters on-demand in cloud computing environments. 

\begin{figure*}[htb]
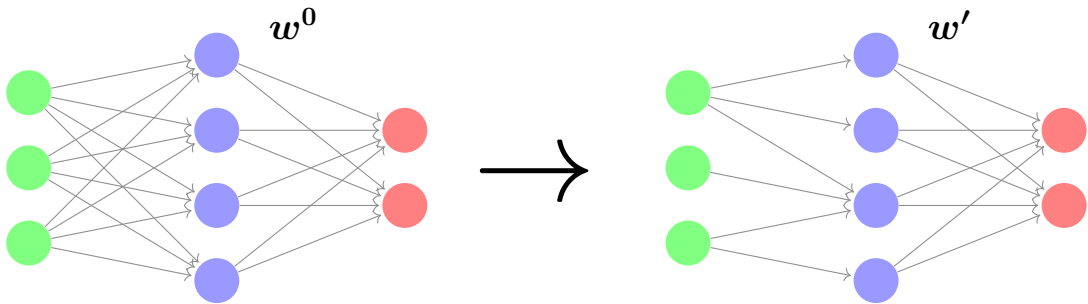

\centering
\begin{neuralnetwork}[height=4]
    \inputlayer[count=3, bias=false]
    \hiddenlayer[count=4, bias=false] \linklayers
    \outputlayer[count=2, bias=false] \linklayers
\end{neuralnetwork}
\put(-60, 100){\Large$\bm{w^0}$}
\hspace{0.5cm}
    \raisebox{4.2em}{\scalebox{5}{$\to$}}
\hspace{0.5cm}
\begin{neuralnetwork}[height=4]
    \inputlayer[count=3, bias=false]
    \link[from layer=0, to layer=1, from node=1, to node=1]
    \link[from layer=0, to layer=1, from node=1, to node=2]
    \link[from layer=0, to layer=1, from node=1, to node=3]
    \link[from layer=0, to layer=1, from node=2, to node=3]
    \link[from layer=0, to layer=1, from node=3, to node=3]
    \link[from layer=0, to layer=1, from node=3, to node=4]
    \hiddenlayer[count=4, bias=false]
    \outputlayer[count=2, bias=false] \linklayers 
\end{neuralnetwork}
\put(-60, 100){\Large$\bm{w'}$}
\caption{Pruning neural network (NN) weights. An example fully connected (feedforward) NN with three input nodes, one hidden layer with four nodes, two output nodes, and weights denoted by $\bm{w^0}$  (left) is pruned, removing half of the weights connecting the input layer with the hidden layer, resulting in a much sparser NN (right) with weights denoted by $\bm{w'}$. The density $d=0.7$ of the pruned model is the ratio between the number of weights in the pruned model (right) divided by the number of weights in the original model (left), and reflects a reduction of 30\% in the number of weights. }
\label{fig:pruning}
\end{figure*}

The computational effort required by these huge DL models is significant, resulting in a high economic cost and power usage for training, storing, and inference (prediction). These factors motivate a recently renewed interest in methods for compressing neural networks (NNs), such as pruning and quantization \cite{cheng2024survey}. In this context, ``pruning'' refers to the removal of weights (also known as connections or edges) in the NN, corresponding to setting a subset of the weights to zero (see \cref{fig:pruning} for an illustrative example), in an effort to reduce the overall memory and inference footprint. This removal of weights can be done before training, during training, or in post-processing (or a combination thereof). 

Pruning methods have been applied to LLMs and have been shown to reduce the model size drastically while resulting in only a small reduction in accuracy (for example, see Refs.~\cite{lagunas2021block, kurtic2022optimal}). 

The intuition behind pruning is that, given a neural network with a large number of parameters, it is likely that the network contains less important connections that can be removed without significantly impacting its performance. This could be explained as a manifestation of the Lottery Ticket Hypothesis  \cite{frankle2018lottery} -- dense trained NNs may contain subnetworks that can achieve the same accuracy in isolation as the full NN. This is particularly clear in cases in which the ML model is overparametrized, meaning that it has more degrees of freedom than the data. In this case, techniques for regularization are commonly used to avoid overfitting. One can think of pruning as a type of regularization, which can actually boost generalization -- by making it more difficult for the NN to overfit the training data. 

Various scalable one-shot (without the need for iterative pruning and fine-tuning) pruning techniques have been used to prune LLMs, such as Wanda and SparseGPT~\cite{sun2023simple, frantar2023sparsegpt}. Our basic premise in this work is that (generally) one can improve upon one-shot pruning techniques by using more in-depth optimization-based pruning techniques, but this can come at a significant computational cost. As a corollary, a feature of this approach is that it allows for a quality-time (or cost) trade-off that is not available when using a one-shot pruning technique alone. 

The main contributions of this paper are in laying out the components of a modular and scalable iterative Combinatorial Brain Surgeon (iCBS):
\begin{itemize}
	\item Solving small problems in each step, limiting the computational requirements for both calculating the Hessian as well as solving the combinatorial optimization problem. The optimization incorporates the updated Hessian and gradient of the respective block at each step. 
	\item Introducing the concept of weight-scoring methods, to be used for initial pruning as well as the selection of weights to be considered in each block. The latter focuses the computational resources where they are best used by selecting the pruned/kept weights that are most likely to be misclassified. 
	\item Employing tabu search to avoid cycling and promote exploration \cite{glover1993user}. 
	\item Limiting the number of weights that must be considered by fixing a large fraction of the weights to be pruned/kept. 
\end{itemize}

The main findings in this work are: 
\begin{itemize}
	\item Our method achieves higher performance metrics at the same density/sparsity levels compared to Wanda~\cite{sun2023simple} and other baseline pruning approaches when applied to large models, including Mistral~\cite{jiang2023mistral} and DeiT~\cite{touvron2021training}.
	\item The observed improvement in performance metrics requires optimization over only a small fraction of the weights, due to fixing weights and effective use of weight-scoring methods for selection of the next block of weights to optimize over. 
	\item The computational bottleneck barring us from optimizing over larger blocks of weights is the block solver (optimizer) run time. Therefore, this method could likely be boosted substantially by the availability and use of hardware accelerators (including quantum and quantum-inspired computers) if they are able to solve quadratic constrained binary optimization problems fast and at scale. Notably, this algorithm is ``quantum-amenable'', meaning that the main building block of our algorithm can be straightforwardly cast as a quantum optimization problem.
\end{itemize}

This paper is structured as follows. In \cref{sec:related_work}, we review related work, and in \cref{sec:preliminaries}, we introduce the problem definition and provide a self-contained derivation of the Combinatorial Brain Surgeon. In \cref{sec:method}, we describe the modified per-block optimization problem, and lay out our proposed algorithm for the iterative Combinatorial Brain Surgeon. In \cref{sec:results}, we present and discuss our results, and in \cref{sec:conclusions}, we present our conclusions and discuss future directions. 

\section{Related work} 
\label{sec:related_work}

The idea of using a Taylor approximation of the loss function for pruning dates back more than 30 years to the Optimal Brain Damage in 1989 \cite{lecun1989optimal}. In this work the Hessian was assumed to be diagonal, so that each weight is pruned without affecting the others. Several years later the Optimal Brain Surgeon was introduced, which takes into account the full Hessian, but prunes one weight at a time \cite{hassibi1993optimal}. Much later, the idea of pruning NNs by carefully balancing the removal of multiple weights was introduced in the Combinatorial Brain Surgeon (CBS) \cite{yu2022combinatorial}.

Over the last few years, other modern pruning techniques have been introduced. Pruning methods can broadly be divided into two categories; simple one-shot methods based on weights, and more sophisticated optimization-based techniques.

\textbf{One-shot Approaches}: The most common and simple pruning methods are variations on zeroing out low-magnitude weights \cite{han2015learning}. These methods assume that removing low-magnitude weights will not change the loss function appreciably. Indeed, in practice typically many weights can be removed in this way without affecting performance significantly \cite{blalock2020state}. Weights and activations (Wanda) is based on a modification to simple magnitude pruning such that weights are pruned if the product of their magnitudes and activations is small \cite{sun2023simple}. It has been applied with some success to LLMs, and we shall refer to it and use it below. 

\textbf{Optimization-based Approaches}: A parallel line of work has centred around using additional information, for example from derivatives of the loss function, to quantify the effect of pruning on the loss function, and then solving an optimization problem to decide which weights to prune. We focus here on methods using the second derivative for this purpose \cite{lecun1989optimal, hassibi1993optimal}. These methods suffer from the need to estimate the Hessian (the matrix containing the second derivatives), which can be computationally expensive, although this can be partially mitigated by simplifying assumptions or by estimating the Hessian using the per-sample gradient \cite{singh2020woodfisher}. Another approach taken, e.g., by CHITA, has been to avoid calculating or estimating the Hessian itself by using only the per-sample gradient \cite{benbaki2023fast}. Regardless, the advantage of this paradigm is that it provides a direct (but approximate) measure of the effect of pruning weights on the loss function. Unfortunately, the requirement of computing the Hessian or other large matrices, in addition to the typical size of the optimization problems involved, have led to difficulties in scaling these approaches, in particular to LLMs. We continue our discussion below by positioning our work as a type of optimization-based pruning.

Block coordinate descent (BCD) is an optimization technique where the objective function is minimized by iteratively optimizing over a subset (block) of the variables, while holding the other variables fixed. The use of BCD in solving large optimization problems is well established in the operations research (OR) community. Within the quantum community, it was introduced for solving large Quadratic Unconstrained Binary Optimization (QUBO) / Ising problems \cite{rosenberg2016building, zintchenko2015local} and popularized via D-Wave's \textsc{qbsolv} solver~\cite{booth2017partitioning}. An important distinction is that in our work the optimization problem solved at each step is a local approximation of the (much more complex) problem to be solved, rather than a sub-QUBO of a larger QUBO. 

\section{Preliminaries}
\label{sec:preliminaries}

In this section we introduce the problem definition, and provide a self-contained derivation of the Combinatorial Brain Surgeon~\cite{yu2022combinatorial} -- the basis for our work.

\subsection{Problem definition}

We start by stating our problem definition in words:

\begin{definition}[\textbf{NN Pruning Problem}]
Given a feature matrix $\bm{X}$, a label vector $\bm{y}$, the real weights of the neural network $\bm{w^0}$ of which there are a total of $N$ non-zero weights, the loss function $\mathcal{L}$, and a target density $d$ (where $d \in (0,1)$), the goal of the Neural Network Pruning Problem is to find the optimum weights $\bm{w'}$ that minimize the value of the loss function while reducing the number of non-zero weights from $N$ to $\lceil dN \rceil$.
\label{def:problem}
\end{definition}

Mathematically, our optimization problem can be stated at a high-level as:
\begin{alignat}{2}
\label{eq:problem_definition}
&{\bm{w'}} = \argmin_{\bm{w}} \, \delta \mathcal{L}(\bm{X}, \bm{y}, \bm{w^0}, \bm{w}) \\
&\text{s.t.} \, \|\bm{w}\|_0 = \lceil d N \rceil , \notag
\end{alignat}
where $\delta \mathcal{L}$ is the change to the loss function due to pruning the weights, $\bm{w^0}$ are the initial weights of the NN (before pruning), $\bm{w}$ are the post-pruning weights of the NN, $\bm{w'}$ are the optimal post-pruning weights, $N$ is the number of weights, $d$ is the target density, and $\|\dots\|_0$ is the $L_0$-norm. The solution of this problem is the collection of optimized (non-zero) weights $\bm{w'}$ that define a NN that has been pruned to the target density $d$. Note that our problem definition is general and flexible, such that it covers most optimization-based pruning techniques and is able to accommodate the different design decisions which are described in detail below. Throughout this work we assume implicitly that the initial weights $\bm{w^0}$ are all non-zero for simplicity of presentation, but initial weights that are zero could be easily accommodated. 

\subsection{The Combinatorial Brain Surgeon}

Much of the work on pruning has focused on the removal of a single weight at a time. A notable downside of this approach is that it does not take into account the interactions between the weights -- it is implicitly assumed that the weights are independent. In contrast, if we take into account the interaction between the weights, we might expect to be able to remove more weights by ``balancing'' (or netting) them out. It turns out that this problem can be formulated as a combinatorial optimization problem known as the Combinatorial Brain Surgeon (CBS) \cite{yu2022combinatorial}. Below we present a self-contained derivation of this optimization problem, before showing how we modified it for our iterative pruner. 

We start by following Refs.~\cite{lecun1989optimal, hassibi1993optimal, kurtic2022optimal} by performing a Taylor expansion of the loss function $\mathcal{L}$ in the change in the weights due to pruning. For the purposes of this derivation, we assume that the NN has already been trained, yielding a vector of $N$ optimized (real) weights $\bm{w^0}$. We define the vector that contains the change in each of the weights due to pruning as $\deltaw \equiv \bm{w} - \bm{w^0}$. 

The Taylor expansion of the loss function $\mathcal{L}$ in $\bm{\delta w}$ around $\bm{w^0}$ gives
\begin{eqnarray}
\label{eq:taylor_expansion}
\mathcal{L}(\bm{w}) &\simeq& \mathcal{L}(\bm{w^0}) + \deltaw^T \nabla \mathcal{L}(\bm{w^0}) + \frac12 \deltaw^T \nabla^2 \mathcal{L}(\bm{w^0}) \deltaw  \notag \\
&+& \mathcal{O}(\deltaw^3).
\end{eqnarray}
At this point, it is common to assume that because the model has been trained, the loss function is at a local minimum, so the gradient is vanishingly small, i.e., $\nabla \mathcal{L}(\bm{w^0}) \simeq \bm{0}$. In addition, it is common to neglect the higher-order terms, i.e., $\mathcal{O}(\deltaw^3)$, due to an assumption that the candidate weights for pruning are all relatively small \cite{yu2022combinatorial} (weight decay can contribute to this too). The assumption that the gradient vanishes is not necessarily true, even for trained models. In fact, we observe that including the gradient yields improved results (as done, e.g., also in Refs.~\cite{singh2020woodfisher, benbaki2023fast}), without incurring significant overhead -- as we shall see, the gradient needs to be calculated anyway to estimate the Hessian. We obtain:
\begin{alignat}{2}
\label{eq:deltaL}
\delta \mathcal{L}(\bm{w}) &\equiv \mathcal{L}(\bm{w})  - \mathcal{L}(\bm{w^0}) \\
&\simeq \deltaw^T \nabla \mathcal{L}(\bm{w^0}) + \frac12 \deltaw^T \nabla^2 \mathcal{L}(\bm{w^0}) \deltaw. \notag
\end{alignat}
We note that $\delta \mathcal{L}$ can, in principle, be evaluated exactly for any candidate vector of weights $\bm{w}$. However, it would be computationally prohibitive to select the optimally pruned weights $\bm{w'}$ by brute-force -- by evaluating $\delta \mathcal{L}$ for all possible values of $\bm{w}$. Exact evaluation is still useful for checking the accuracy of the assumptions and tracking progress, such as in the context of an algorithm involving multiple updates (like our algorithm, see discussion below). 

We define the Hessian ${\mathcal{H}^0 \equiv \nabla^2 \mathcal{L}(\bm{w^0})}$ and the gradient ${\mathcal{G}^0 \equiv \nabla \mathcal{L}(\bm{w^0})}$ of the loss function at the initial weights $\bm{w^0}$. Then we can write the optimization problem in our problem definition \cref{eq:problem_definition} explicitly using \cref{eq:deltaL} as:
\begin{alignat}{2}
\label{eq:problem_definition_explicit}
&\argmin_{\{ \delta w_i \}} \, \alpha \sum_i \delta w_i \mathcal{G}^0_i + \frac12 \sum_{i,j} \delta w_i \mathcal{H}^0_{i,j} \delta w_j + \lambda \sum_i (\delta w_i)^2 \\
&\text{s.t.} \quad \forall i: \, \sum_{i} \mathds{1}[w_{i}\neq 0] = \lceil d N \rceil \quad \text{and} \quad \delta w_i \equiv w_i - w^0_i \notag
\end{alignat} 
where $\alpha \geq 0$ is a coefficient that controls the relative importance of the gradient term versus the Hessian term (see Ref.~\cite{benbaki2023fast} for a discussion), and $\mathds{1}[\dots]$ is the indicator function. We added the ridge term here to control the change in the weight vector, i.e, $\bm{\delta w}$, which may be useful since we expect the Taylor approximation to break down if the change is large (see Ref.~\cite{benbaki2023fast} for a discussion). Note that from here onward we switch from vector/matrix notation to element notation, which is more verbose but arguably clearer (and more customary) in the definition of optimization problems. 

We now follow Ref.~\cite{yu2022combinatorial} to define binary variables ${x_i \in \{0,1\}}$ which indicate whether weight $i$ will be pruned (in which case, $x_i=1$), i.e., ${w_i=(1-x_i)w^0_i}$. Initially, before pruning, all the weights are set to non-zero values, so ${x_i=0}$ for all $i$ and ${w_i=w^0_i}$ and therefore ${\delta w_i = 0}$. We note that
\begin{equation}
\label{eq:delta_w}
\delta w_i \equiv w_i - w^0_i = (1-x_i)w^0_i - w^0_i = -x_i w^0_i.
\end{equation}
This allows us to formulate an optimization problem to determine which weights should be pruned, i.e., to find the optimized $x_i$ that minimize the change in the loss function. Accordingly, we substitute this definition into \cref{eq:problem_definition_explicit} and obtain the following quadratic constrained binary optimization (QCBO) problem
\begin{alignat}{2}
\label{eq:qcbo}
    &\!\min      \quad && - \alpha \sum_{i=1}^N \left( w^0_i \mathcal{G}^0_i \right) x_i + \frac12 \sum_{i,j=1}^N \left( w^0_i \mathcal{H}^0_{ij}w^0_j \right) x_i x_j \\
    & && + \lambda \sum_{i=1}^N x_i (w^0_i)^2 \notag \\
    & \text{s.t.} \quad && \sum_{i=1}^N x_i = \lceil (1-d)N \rceil \notag \\
    &             \quad && x_i \in \{0, 1\} \quad \forall i \in \{1, \dots, N\}, \notag
\end{alignat}
where we used the fact that ${x_{i}^2=x_{i}}$ for ${x_{i} \in \{0, 1\}}$. This optimization problem can, in principle, be solved to obtain a pruned model. Solving it for a range of densities $d$ would allow one to explore the sparsity-accuracy trade-off. However, the issue is that for DL models this optimization problem is generally intractable, because the number of weights $N$ is large (e.g., up to ${\sim}10^{11}$ in recently published LLMs \cite{dubey2024llama}). Even constructing the Hessian, which is required in order to solve this problem, can be prohibitive. Once the Hessian is constructed, the second hurdle is the number of variables in this problem, because there is a variable for each weight, and the number of weights $N$ is typically very large. So, noting that this formulation is generally not scalable, we discuss how an iterative approach can utilize this same logic while scaling to large models in the next two sections.

Our derivation above is based on Ref.~\cite{yu2022combinatorial} but differs from it in the addition of the gradient and ridge terms. In addition, in that work the authors formulated also a second optimization problem in which the loss approximation was used to fine-tune the weights (continuously). In our work we only discuss how to solve the pruning problem, leaving fine-tuning in this way (and in general) to future work. 

We remark that the optimization problem \cref{eq:qcbo} is ``quantum-amenable'', in that it could, in principle, be solved by a quantum computer. Firstly, such optimization problems can be reformulated as a QUBO problem by adding a penalty term that encodes the single constraint. There are several quantum algorithms for solving QUBO problems, for example the Quantum Approximate Optimization Algorithm (QAOA) \cite{farhi2014quantum}. There are also algorithms that aim to incorporate constraints in a more native way, such as via constraint-preserving mixers \cite{hadfield2017quantum}, or using quantum constrained Hamiltonian optimization techniques (Q-CHOP) \cite{perlin2024q}. 

\section{Method}
\label{sec:method}

\subsection{The per-block formulation}

We propose, devise, and implement an algorithm that solves an optimization problem similar to \cref{eq:qcbo} iteratively at each step, for a \textit{block} of $n$ weights (typically, $n \ll N$). To this end, we maintain a vector $\bm{w^c}$ with the \textit{current} state of the weights, given the pruning decisions made in the steps completed so far (in the first step, we have $\bm{w^c}=\bm{w^0}$). Notice that our discussion and equations in the previous section referred only to the initial weights $\bm{w^0}$ and the final (optimized/pruned) weights $\bm{w}$. We now revisit the above logic while incorporating the iterative updating of the weights. Instead of \cref{eq:delta_w} we now have:
\begin{equation}
\label{eq:delta_w_iterative}
\delta w_i \equiv w_i - w^c_i = (1-x_i)w^0_i - w^c_i = - \left( \Delta w_i + x_i w_i^0 \right), 
\end{equation}
where for convenience we have defined the change from the initial weights to the current weights as ${\Delta w_i \equiv w^c_i - w^0_i}$. This is convenient because it allows us to separate the part of this equation that depends on the decision variables $x_i$ from the part that does not. Note that this equation refers to both the initial weights $w^0_i$ and the current weights $w^c_i$, unlike \cref{eq:delta_w}.

We define the block Hessian $\mathcal{H}^B \equiv \nabla^2 \mathcal{L}(\bm{w^c})$ and the block gradient $\mathcal{G}^B \equiv \nabla \mathcal{L}(\bm{w^c})$ which are now calculated at the current weights $\bm{w^c}$ (rather than the initial weights $\bm{w^0}$). By substituting \cref{eq:delta_w_iterative} into \cref{eq:problem_definition_explicit} we arrive at the following block QCBO problem:
\begin{alignat}{2}
\label{eq:qcbo_block}
    &\!\min \,                   && - \alpha  \sum_{i=1}^n \left(w^0_i \mathcal{G}^B_i \right) x_i + \sum_{i,j=1}^n \left(w_i^0 \mathcal{H}_{ij}^B \Delta w_j \right) x_i \\
    &\phantom{\!\min} \, && + \lambda \sum_{i=1}^n \left[2 \Delta w_i w_i^0 + (w_i^0)^2 \right] x_i \notag \\
    &                              && + \frac12 \sum_{i,j=1}^n \left( w_i^0 \mathcal{H}_{ij}^B w_j^0 \right) x_i x_j \notag \\
    & \text{s.t.} \,            && \sum_{i=1}^n x_i = k \notag \\
    &                \,            && x_i \in \{0, 1\} \quad \forall i \in \{1, \dots, n\}, \notag
\end{alignat}
where we have discarded an irrelevant constant in the objective, and $k \in (0, n)$ is the number of weights to be pruned in this block. This allows us to devise an algorithm that relies on solving a series of such problems, updating a block of weights at each step, as we describe in the following. Note that \cref{eq:qcbo} can be recovered by considering the case in which the block includes all the weights, i.e., $n = N$, the current weights are identical to the initial weights, i.e., $w^c_i=w^0_i$, and therefore $\Delta w_i = 0$ for all $i$, and ${\mathcal{G}^B = \mathcal{G}^0}$ and ${\mathcal{H}^B = \mathcal{H}^0}$. 

This optimization problem requires the Hessian and the gradient for the chosen block of variables to be calculated. The Hessian can be calculated by taking the gradient of the gradient; however, this can be prohibitively computationally expensive. As shown in Ref.~\cite{hassibi1993optimalextensions} if one calculates the matrix of per-sample gradients $\mathcal{A}$ for the given block, the block Hessian can be estimated more cheaply via
\begin{equation}
\label{eq:hessian_estimate}
\mathcal{H}^B \simeq \frac1n \mathcal{A}^T \mathcal{A},
\end{equation}
where $n$ is the number of samples in the respective batch, and that is the approach we use in this work. 

\subsection{The iterative Combinatorial Brain Surgeon}

\begin{figure*}[htb]
  		\centering
		\includegraphics[width=0.8\textwidth]{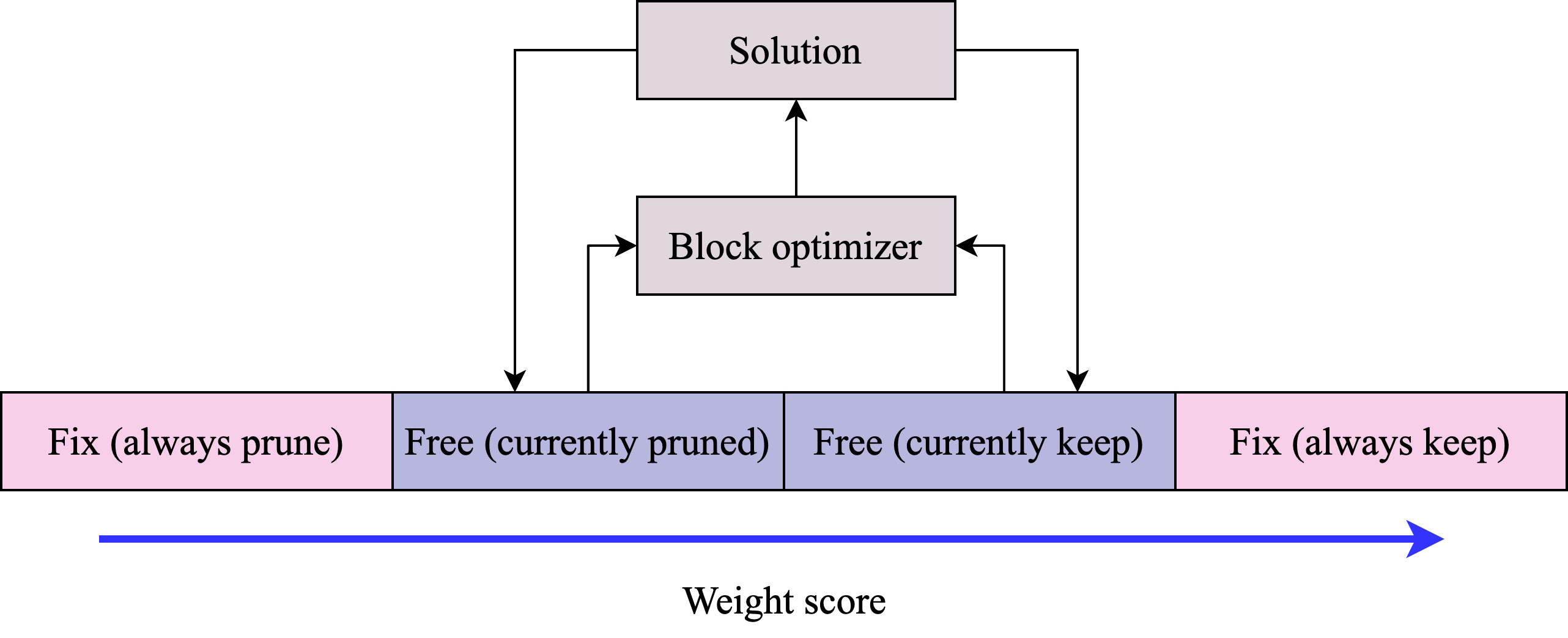}
  \caption{Schematic illustration of the per-block pruning process in iCBS. The \codevar{init\_method} weight-scoring method is used to score all the weights at the beginning. Weights with extreme scores are fixed to always be pruned/kept, meaning that these weights are taken out of the candidate pool for the per-block optimization. Then, in each step, the \codevar{selection\_method} weight-scoring method is used to score the non-fixed (free) weights. Then, the weights to be optimized over are selected from the currently pruned/kept sets, taking into account which weights are tabu-ed. Then the gradient and Hessian are estimated for that block, an optimization problem is constructed and passed to the block optimizer. The block optimizer solves the per-block optimization problem and returns a solution. Finally, the currently kept / pruned sets are updated and the tabu list is updated (not pictured).}
  \label{fig:icbs_diagram}
\end{figure*}

We now describe in detail all the steps involved in our pruning algorithm (see the pseudo code \cref{algo:pruner} and the schematic diagram \cref{fig:icbs_diagram}) and the rationale behind them. We refer to this algorithm as ``iterative Combinatorial Brain Surgeon'' (iCBS). 

{\bf Initial solution} -- The initial solution is an initial decision of which weights should be pruned based on the \codevar{init\_method} weight-scoring method. Here we introduce the term ``weight-scoring method'' to refer to any one-shot metric-based method of assigning a score to each weight (here a total of $N$ scores). Then we prune the $\lceil (1-d)N \rceil$ weights with the lowest scores. For example, in the magnitude weight-scoring method we simply prune the lowest-magnitude weights. See \cref{tab:weight_scoring_methods} for a summary of the weight-scoring methods included in this work. Following Ref.~\cite{sun2023simple} we allow the weight-scoring methods (except for ``Random'') to be applied on a per-layer basis, per-output basis, or per-input basis. This is based on the observation in Ref.~\cite{sun2023simple} that for LLMs the per-output version of Wanda performed better. The Wanda and Gradient weight-scoring methods are calculated over the ``calibration data'' of size \codevar{batch\_size\_calibration}, which is typically larger than the pruning batch size.
%
\begin{table}[htb]
    \centering
 	\caption{Definition of weight-scoring methods included in this work. $w_i$ is the $i$-th weight, $\mathcal{G}_i$ is its gradient, and $a_i$ is its activation, averaged across the batch.}
	\begin{tabularx}{0.7\columnwidth}{@{\hskip 0.45in}@{}l@{\hskip 0.45in}l@{}}
		\toprule
        \toprule
   		Name        & Score \\
		\midrule
		\texttt{Random}           & Random  \\ 
		\texttt{Magnitude}        & $| w_i |$  \\ 
		\texttt{Gradient}           & $ | w_i \mathcal{G}_i |$ \\	
		\texttt{Wanda}             & $| w_i a_i^2 |$  \\ 
		\bottomrule
        \bottomrule
	\end{tabularx}
	\label{tab:weight_scoring_methods}
\end{table}

{\bf Fixing weights} -- Given a weight-scoring method, we fix a fraction of the weights with the most extreme scores (high or low). More specifically, the \codevar{fix\_frac\_prune} weights with the lowest scores in the initially pruned set are fixed to always be pruned. Similarly, the \codevar{fix\_frac\_keep} weights with the highest scores in the initially kept set are fixed to always be kept. This is based on the idea that extreme weight scores indicate that those weights should be pruned/kept with high confidence, without requiring the careful balancing that occurs when solving the per-block optimization problem (see \cref{fig:icbs_diagram}). As an example, when using the option of fixing weights with magnitude-based pruning, the weights with the smallest magnitudes are always pruned -- there is not much use in trying to carefully balance their removal. This step is useful in decreasing the number of candidate weights that are considered for per-block pruning in each step (as detailed below). 

{\bf Epochs and steps} -- The pruner performs \codevar{num\_steps} steps for each epoch of \codevar{num\_epochs}. Each epoch consists of a series of per-block optimization steps with an evaluation on the validation data at the end. For each step a random batch of data $X$ and respective labels $y$ (both of length \codevar{batch\_size\_pruning}) are drawn from the training data. In addition, the layer to be pruned is chosen randomly, such that the number of steps per layer in each epoch is roughly proportional to the base-10 log of the number of weights in each layer. This is done in order to assign more steps to larger layers, while still leaving some steps to be assigned to smaller layers.

{\bf Selection} -- In each step we select $n$ weights to be optimized over from the candidate set. The selection is based on the \codevar{selection\_method} weight-scoring method: We select the $k$ highest-scoring candidate weights from the currently pruned set (most likely to be mistakenly pruned) and the $n-k$ lowest-scoring candidate weights from the currently kept set (most likely to be mistakenly kept), in an effort to focus the optimization on weights that are most likely to be currently misclassified. This rationale can be viewed as being inspired by extremal optimization, an optimization algorithm in which the part of the solution with the lowest fitness is updated in each step \cite{boettcher2004extremal}. Without loss of generality, we set $k=\lceil n/2 \rceil$. The candidate set for selection includes only weights that were not fixed (see above), and that are not in the tabu list (see below). 

{\bf Tabu list} -- In order to avoid selecting the same weights repeatedly, we implement a tabu list for each layer \cite{glover1993user}. In each step, when the $n$ weights for the block are selected, the weights in the respective tabu list are excluded from the candidate list. After the selection, the $n$ weights that are selected are added to the tabu list for the respective layer. The maximum length of each tabu list is \codevar{tabu\_frac} of the number of weights in that layer. 

{\bf Estimate the gradient and Hessian} -- The mean gradient is estimated based on the current batch. The Hessian is estimated based on the per-sample gradient via \cref{eq:hessian_estimate}. The calculation of these gradients can typically be done much faster on a GPU than on a CPU.

{\bf Construct the per-block optimization problem} -- The optimization problem is constructed based on \cref{eq:qcbo_block}. Because the coefficients in the problem are often tiny, it is useful to scale them up. In this case we choose a scaling factor such that the mean of the absolute value of the non-negative coefficients is one. Then, we set to zero any elements with an absolute value less than or equal to $10^{-12}$, as these negligibly small values cannot meaningfully contribute to the solution.

{\bf Solve the per-block optimization problem} -- Our pruner is modular and can utilize a range of QCBO and QUBO block solvers to solve the optimization problem. In this study, we use a constrained simulated annealing solver which performs \codevar{num\_restarts} independent starts in parallel, one on each CPU \cite{haener2024solving}. This solver solves this QCBO problem \textit{natively}, meaning that it searches only the feasible space. 

{\bf Apply the block's solution to the model} -- A solution to the above optimization problem is a binary vector $\bm{x}$ indicating which $k$ of the $n$ weights in the selected block should be pruned. Applying the solution requires pruning those $k$ weights as well as un-pruning the $n-k$ remaining weights in the block. 

{\bf Calculate loss and accuracy on validation data} -- In order to evaluate the effect of pruning on the model's performance, at the end of each epoch the loss and accuracy are calculated over the validation data.

\begin{figure*}
\begin{center}
\noindent\begin{minipage}{1.0\linewidth}
\lstset{caption={Pseudo code for the iterative Combinatorial Brain Surgeon (iCBS) algorithm}}
\begin{lstlisting}[label={algo:pruner}] 
Construct initial solution based on init_method weight-scoring method
Fix weights based on fix_frac_prune and fix_frac_keep for all layers
For each epoch of num_epochs:
    For each step of num_steps_per_epoch and batch of data X, labels y, and randomly chosen layer:
        Select n candidate weights based on selection_method and tabu list
        Estimate gradient
        Estimate Hessian (only the elements needed)
        Construct the per-block optimization problem
        Solve the optimization with the block_solver to choose k weights to prune (out of n)
        Apply the block's solution to the model -- pruning or un-pruning weights, as needed
        Add the selected weights to the tabu list for this layer
    Calculate loss and/or accuracy on validation data
\end{lstlisting}
\end{minipage}
\end{center}
\end{figure*}

\section{Benchmarking methodology and results}
\label{sec:results}

We consider the following research questions (RQ) to demonstrate the effectiveness of our approach: 

\begin{questions}[itemindent=1em]
\item Is iCBS able to effectively prune ML models, compared with common baselines?
\item Can iCBS scale to large models, including LLMs? 
\end{questions}

\subsection{Benchmarking methodology}

{\bf Models and datasets} -- The models and datasets used in our work are described in \cref{tab:models_and_datasets}. The Garment Classifier is an example of a simple feedforward network used to classify images of garments into one of ten classes. The DeiT model is a larger Transformer~\cite{vaswani2017attention} model that is used to classify images into one of 1000 classes. Finally, the Mistral-7b model is an LLM that is a foundation model, i.e., it can be used for a range of language-related tasks. Note that we only prune linear layers, excluding other types of layers such as batch and layer normalization layers. In addition, we follow other work by not pruning the final layer if it is found to be advantageous to do so (this was the case for DeiT and Mistral, but not for the Garment Classifier).

\begin{table*}[htb]
	\centering
 	\caption{Models, datasets, and relevant citations. ``Model'' is the model name, ``Weights'' is the total number of weights, ``Tensors'' is the number of weight tensors (connecting two layers) to be pruned, ``Dataset'' is the name of the dataset used for the pruning data batches, ``Train'' is the number of samples in the training set, ``Evaluation'' is the method of evaluation, ``Classes'' is the number of classes for evaluation via classification accuracy, and ``Valid'' is the number of samples in the validation set. The Garment Classifier model is a simple linear model we created with two hidden layers with 512 nodes -- the other models are well known. When the evaluation method was the LM Evaluation Harness, the evaluation was based on the mean zero-shot accuracy across the following seven tasks: BoolQ \cite{clark2019boolq}, RTE \cite{wang2018glue}, HellaSwag \cite{zellers2019hellaswag}, WinoGrande \cite{sakaguchi2021winogrande}, ARC Easy and Challenge \cite{clark2018think}, and OpenbookQA \cite{mihaylov2018can}. The number of classes and samples in the validation set differs across these tasks (hence the asterisk (*) in the table).}
	\begin{tabular}{@{}l@{\hskip 0.2in}r@{\hskip 0.2in}r@{\hskip 0.2in}l@{\hskip 0.2in}l@{\hskip 0.2in}l@{\hskip 0.2in}l@{\hskip 0.2in}l@{}}
		\toprule
        \toprule
   		Model                             & Weights & Tensors & Dataset            & Train & Evaluation & Classes & Valid                                                                                       \\
		\midrule
		Garment Classifer                 &  669K &          3 & Fashion-MNIST \cite{xiao2017fashion}  & 60K   & Classification Accuracy & 10 & 10k  \\ 
		DeiT \cite{touvron2021training}   &    86M &        72 & ImageNet-1K \cite{deng2009imagenet} &  1.3M & Classification Accuracy  &  1k & 50k  \\ 
		Mistral-7b \cite{jiang2023mistral} &     7B &      224 & C4-en \cite{raffel2020exploring}             & 364M & LM Evaluation Harness \cite{eval-harness} & * & * \\
        \bottomrule
        \bottomrule
	\end{tabular}
	\label{tab:models_and_datasets}
\end{table*}

{\bf Parameters} -- We performed hyper-parameter tuning with up to 100 configurations using \textsc{Ray Tune} \cite{liaw2018tune} with \textsc{Hyperopt} \cite{bergstra2013making}, for the Garment Classifier and DeiT models. The parameter values used were obtained by examining the approximately 20 best configurations and averaging over the best values for continuous parameters and choosing the most frequently occurring value for the discrete parameters. For the Mistral model it was not feasible to do large-scale hyper-parameter tuning due to the costs associated with these experiments. The parameter values used, a description of the parameters, and some information on how they were selected are provided in \cref{sec:parameters}.

{\bf Computational resources} -- All benchmarking experiments were run on Amazon Elastic Compute Cloud (EC2) instances. The Garment Classifier and DeiT model pruning experiments were performed on a single \texttt{g5.4xlarge} instance, which has 16 vCPUs, 64 GiB of RAM, and one NVIDIA A10G Tensor Core GPU with 24 GiB of memory. The Mistral model pruning experiments were performed on multiple \texttt{g5.48xlarge} instances, which have 192 vCPUs, 768 GiB of RAM, and 8 NVIDIA A10G GPUs with 24 GiB of memory each.

{\bf Baselines} -- We compare our pruning method iCBS to two commonly-used one-shot pruning methods, namely Wanda~\cite{sun2023simple} and magnitude pruning \cite{han2015learning}. We also include a gradient-based one-shot pruning method (simply dubbed Gradient) that is motivated by pruning weights that have a small first-order contribution (see \cref{eq:deltaL} and \cref{eq:qcbo_block}) to the change in the loss function. For each of these three baselines we tested three variations---aggregation per-layer, per-input, and per-output---and include in the figures the variation that had the highest accuracy values. Finally, we include random pruning to a target density as a lower baseline. The best baseline was also used for the initial pruning of iCBS. 

\subsection{Results and Discussion}

\textbf{Effective pruning (RQ1) -- Garment Classifier} -- To establish whether iCBS can effectively prune NNs, we start with a relatively small experiment using a simple NN with two hidden layers, and therefore three weight tensors (which connect the layers) to be pruned. We refer to this model as the ``Garment Classifier''. The task is the classification of $28 \times 28$ grayscale images of garments accurately, out of 10 classes (see the Fashion-MNIST dataset \cite{xiao2017fashion}). We trained the Garment Classifier for 100 epochs with the Stochastic Gradient Descent (SGD) optimizer in \textsc{PyTorch} \cite{paszke2019pytorch}. 

\begin{figure}[htb]
\includegraphics[width=1.0 \columnwidth]{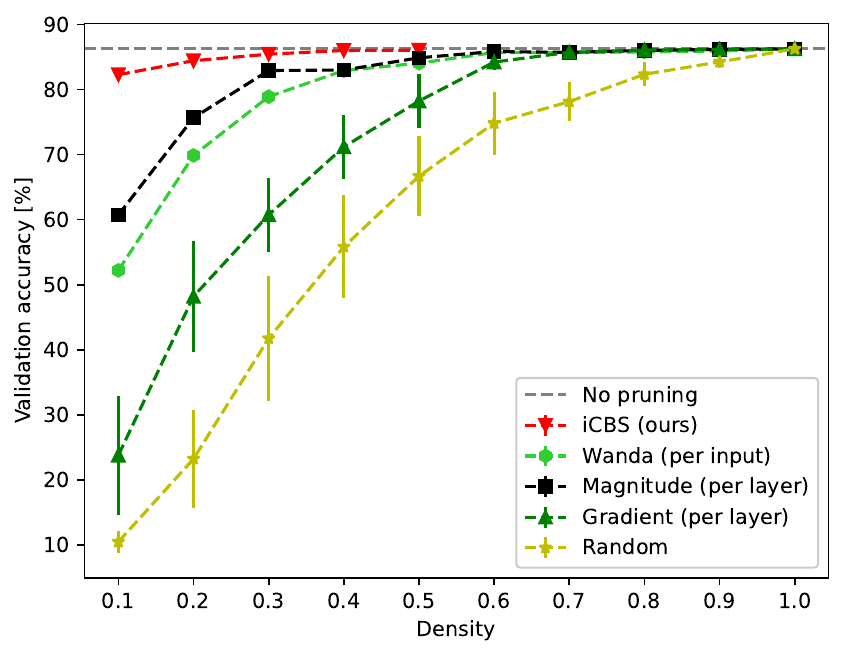}
\caption{Results for the Garment Classifier model on the Fashion-MNIST dataset. This plot shows the dependence of the final (post-pruning) validation accuracy (top-1) on the density for various types of pruning -- the baselines and our pruner iCBS. The horizontal line labeled ``No pruning'' shows the validation accuracy of the unpruned model. Error bars are included for all baselines except magnitude (since it is deterministic) and show the standard deviation over 30 random repetitions.}
\label{fig:garment_classifier_accuracy_vs_density}
\end{figure}

Our pruning results are presented in \cref{fig:garment_classifier_accuracy_vs_density}. In this case, the magnitude-based pruning was the best performing baseline, unlike for the other models used in this work, for which Wanda was the best baseline. This may be related to the other models being Transformers, whereas the Garment Classifier is a simple feedforward NN. We see that a significant portion of the weights (${40\mbox{ -- }50\%}$) can be pruned using magnitude-based pruning with a negligible change in validation accuracy. The \textit{room for improvement}, i.e., the difference between the initial pruning by the per-layer magnitude-based method and the ``No pruning'' horizontal line (the performance of the unpruned model) is largest at small densities, and declines until it vanishes (around density 60\%).

Our pruner, iCBS, starts out with initial pruning using magnitude-based pruning. As such, the difference between the per-layer magnitude curve and the iCBS curve can be attributed to our pruner. Overall, the improvement is largest (+21.6\% in accuracy, at density 10\%) at low densities and declines as the density is increased and the room for improvement decreases and eventually vanishes. When using iCBS to prune this model, even at a low density of 10\%, the final validation accuracy only decreased by a few percent versus the unpruned model (from 86.2\% to 82.3\%). The iCBS pruning was done over \codevar{num\_steps}~$=300$ steps for each epoch of \codevar{num\_epochs}~$=10$ epochs, with a block size of $n=1024$. 

\begin{figure*}[htb]
  \begin{minipage}{\textwidth}
	\begin{subfigure}[t]{0.48\textwidth}
	        \centering
		\includegraphics[width=\textwidth]{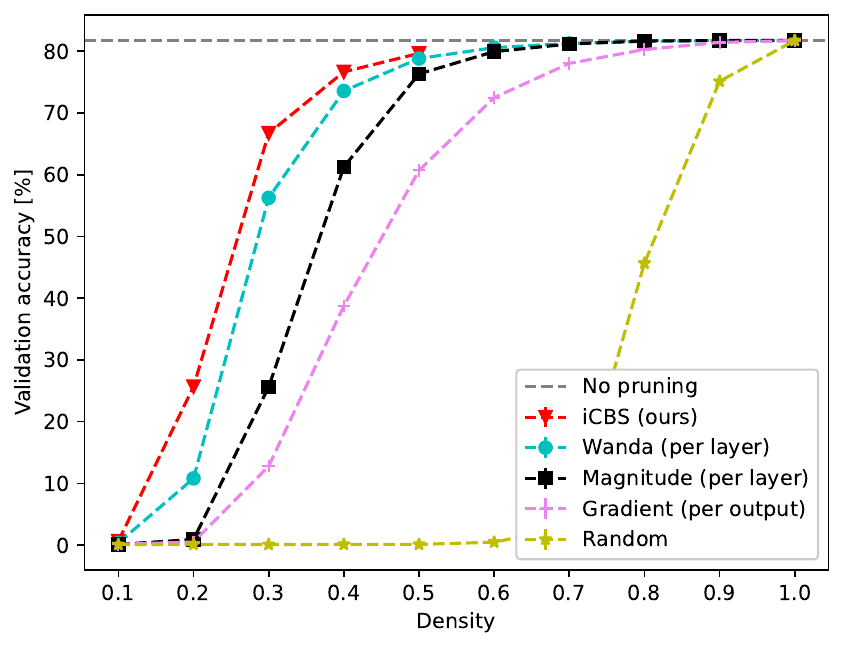}
		\caption{Validation accuracy vs. density}\label{fig:deit_accuracy_vs_density}
	\end{subfigure}\hspace{0.5cm}
	\begin{subfigure}[t]{0.48\textwidth}
		\centering
		\includegraphics[width=\textwidth]{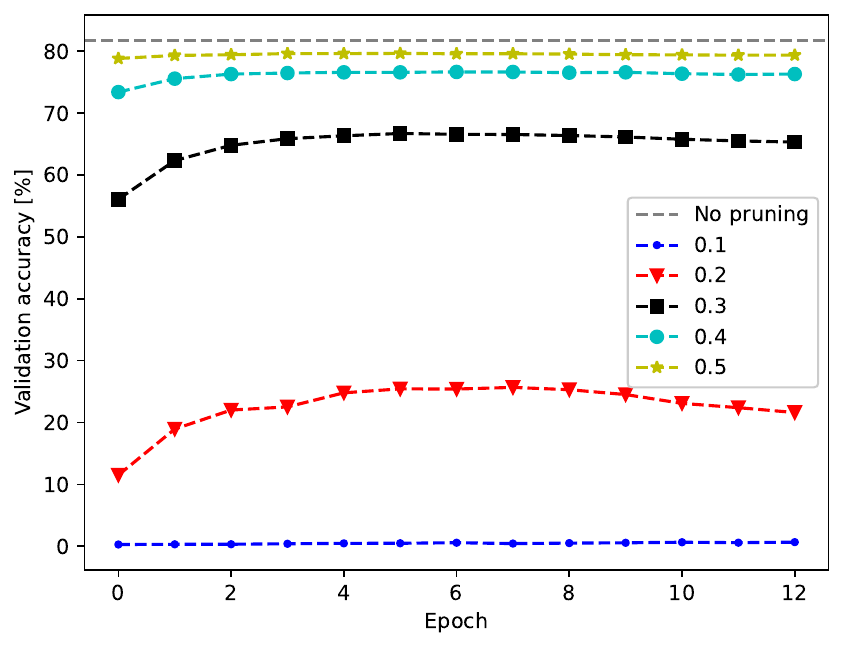}
		\caption{Validation accuracy vs. epoch for iCBS}\label{fig:deit_evolution}
	\end{subfigure}
  \end{minipage}
  \caption{Results for the DeiT model on the ImageNet-1K dataset. These plots shows the dependence of the post-pruning validation accuracy (top-1) on the density for various types of pruning -- the baselines and our pruner iCBS. The horizontal line labeled ``No pruning'' shows the validation accuracy of the unpruned model.}
  \label{fig:deit}
\end{figure*}

\textbf{Scaling (RQ2) -- DeiT} -- Encouraged by these results, we now explore whether iCBS would also perform well for larger models and data sets. Our first larger experiment was on the DeiT (Data Efficient Image Transformer) model, a Vision Transformer (ViT) that is trained efficiently \cite{touvron2021training}. The task is the classification of images accurately, out of 1000 classes (see the ImageNet-1K dataset \cite{deng2009imagenet}). The images are of various sizes, and are commonly centered and cropped -- DeiT center-crops them to $224 \times 224$ pixels. We used the pre-trained DeiT-base model, obtained from \textsc{HuggingFace}. 

Our pruning results are presented in \cref{fig:deit}. In \cref{fig:deit_accuracy_vs_density} we see that the per-layer Wanda baseline is the most performant with per-layer magnitude being close behind. We see that a significant portion of the weights ({30 -- 40\%}, slightly less than for the Garment Classifier) can be pruned by either of these methods with a negligible change to the validation accuracy. Unlike for the Garment Classifier, here all methods led to zero accuracy at density 10\%, presumably due to this task being significantly harder, and perhaps due to less redundancy in the model. 

Our pruner, iCBS, starts out with initial pruning using the Wanda layer-based pruning. As such, the difference between the per-layer Wanda curve and the iCBS curve can be attributed to our pruner. iCBS beat all the baselines for all densities with room for improvement. Overall, the improvement is largest at low densities (+14.2\% and +10.6\% in accuracy, at densities 20\% and 30\%, respectively) and declines as the density is increased and the room for improvement decreases and eventually vanishes. The iCBS pruning was done over ${\codevar{num\_steps}=500}$ steps for each epoch of ${\codevar{num\_epochs}=10}$ epochs, with a block size of $n=1024$. In \cref{fig:deit_evolution} we see that for low and high densities the validation accuracy converged quickly. For the middle densities ${20\mbox{ -- }30\%}$ convergence was slower.

\textbf{Scaling (RQ2) -- Mistral-7b} -- Our second larger experiment was on the Mistral model \cite{jiang2023mistral}. This model was pruned using the C4 dataset \cite{raffel2020exploring}, and evaluated using the LM Evaluation Harness \cite{eval-harness} on seven tasks (as done in Ref.~\cite{sun2023simple}) listed in the caption of \cref{tab:models_and_datasets}. We used the pre-trained Mistral-7b model (version 0.1), obtained from \textsc{HuggingFace}. To reduce GPU memory requirements, we loaded this model using float16 and a shortened context length of 4096. 

\begin{figure*}[htb]
  \begin{minipage}{\textwidth}
	\begin{subfigure}[t]{0.48\textwidth}
	        \centering
		\includegraphics[width=\textwidth]{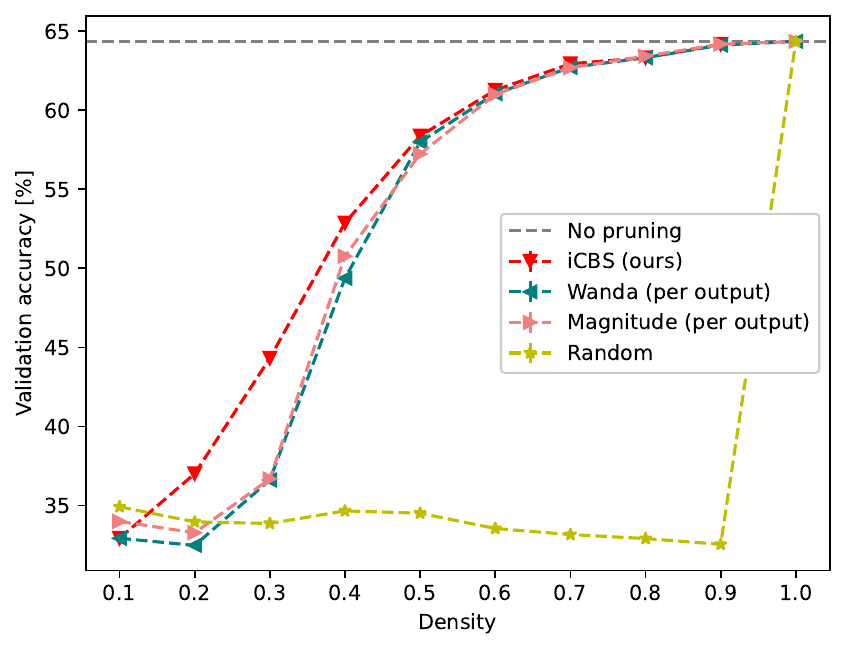}
		\caption{Accuracy vs. density}\label{fig:mistral_7b_accuracy_vs_density}
	\end{subfigure}\hspace{0.5cm}
	\begin{subfigure}[t]{0.48\textwidth}
		\centering
		\includegraphics[width=\textwidth]{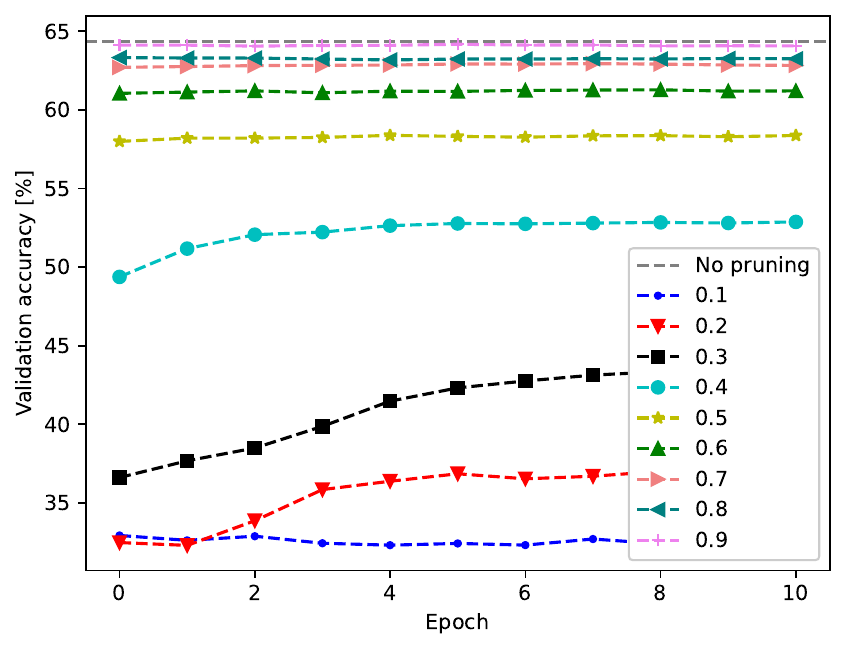}
		\caption{Validation accuracy vs. epoch for iCBS}\label{fig:mistral_7b_evolution}
	\end{subfigure}
  \end{minipage}
  \caption{Results for the Mistral-7b model pruned using the C4 dataset and validated using the LM Evaluation Harness. These plots shows the dependence of the post-pruning validation accuracy (top-1) on the density for various types of pruning -- the baselines and our pruner iCBS. The horizontal line labeled ``No pruning'' shows the validation accuracy of the unpruned model.}
  \label{fig:mistral_7b}
\end{figure*}

Our pruning results are presented in \cref{fig:mistral_7b}. In \cref{fig:mistral_7b_accuracy_vs_density} we see that the per-output Wanda baseline performed similarly to per-output magnitude, unlike in Ref.~\cite{sun2023simple} where the former was shown to outperform the latter for other LLMs (which we have verified). In this case, only a smaller portion of the weights ${\sim}10\%$ could be pruned by either of these methods with a negligible change to the validation accuracy. Unlike for DeiT, at the lowest density of 10\% the validation accuracy is not zero. The reason is that the tasks the model was evaluated on are yes/no or multiple-choice questions, such that even a completely random model would be expected to achieve a finite accuracy of $\simeq 35\%$ (by random guessing). This also explains why the random pruning baseline was able to match the performance of all the other pruning methods we evaluated. 

Our pruner, iCBS, starts with initial pruning using the Wanda per-output pruning. Therefore, the difference between the per-output Wanda curve and the iCBS curve can be attributed to our pruner. The improvement is largest (+7.7\% in accuracy, at density 30\%) at low densities and declines as the density is increased and the room for improvement decreases and eventually vanishes. The iCBS pruning was done over ${\codevar{num\_steps}=300}$ steps for each epoch of ${\codevar{num\_epochs}=10}$ epochs, with a block size of ${n=4096}$. In \cref{fig:mistral_7b_evolution} we observe that for low and high densities the validation accuracy converged quickly. For the middle densities ${20\mbox{ -- }40\%}$, convergence was slower. 

\begin{figure}[htb]
\includegraphics[width=1.0 \columnwidth]{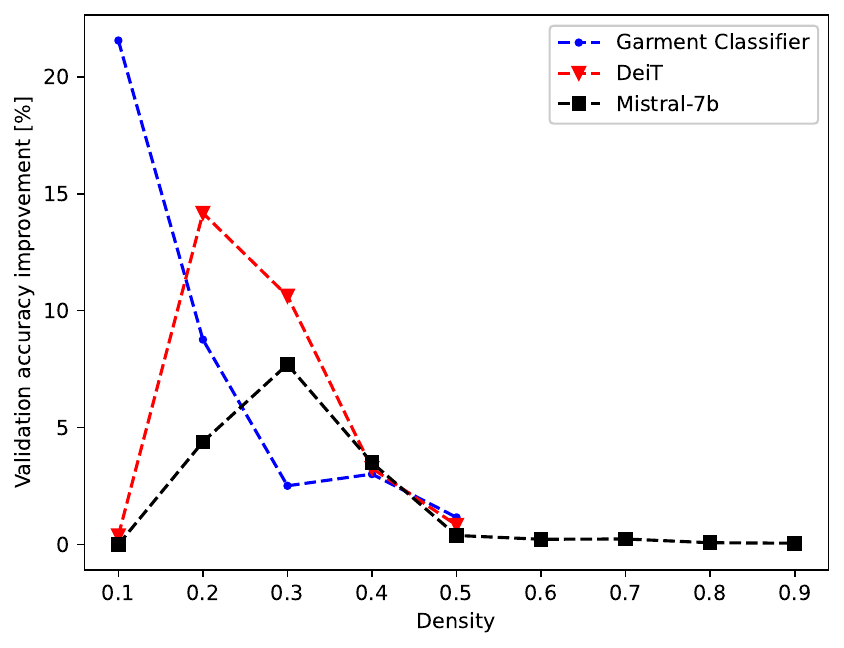}
\caption{Improvement in validation accuracy of iCBS over the initial pruning vs. the density, for each of the models considered.}
\label{fig:improvement}
\end{figure}

\textbf{Summary of improvement} -- We summarize the improvement (in validation accuracy) of iCBS over the respective baseline (used for initial pruning) in \cref{fig:improvement}. For all of the models the improvement drops off for high densities. This is partially because there is not much room for improvement (at very high densities), i.e., the best baseline is able to match the unpruned model's performance. In addition, we observe that iCBS is able to improve on the baselines most for densities around ${20\mbox{ -- }40\%}$. In some cases, it cannot improve on the baseline despite significant room for improvement, such as for Mistral-7b at 50\% density. 

\textbf{Fraction of weights optimized} -- Recall that in each epoch of \codevar{num\_epochs} the pruner performs \codevar{num\_steps} and at each step the pruner optimizes over a block of size $n$ of weights. By multiplying the total number of steps and the block size, we can find an upper bound on the number of weights optimized. If we divide this number by the total number of weights (and truncate to one if it is above), we find the fraction of weights optimized. See \cref{tab:fraction_optimized} for the fraction of weights optimized in each of our three experiments. This highlights the fact that for DeiT and Mistral-7b only a small fraction of weights was optimized over, while still achieving notable validation accuracy improvements, as seen in \cref{fig:improvement}. We suggest that contributing reasons to this are the fixing of weights and the effective use of weight-scoring methods for selection of the next block of weights to optimize over. 

\begin{table}[htb]
	\centering
 	\caption{Fraction of weights optimized. ``Model'' is the model name, ``Weights'' is the total number of weights, ``Optimized'' is the number of weights optimized, calculated by multiplying the total number of steps and the block size $n$, and ``Fraction'' is the fraction of the weights that were optimized over (max 1).}
	\begin{tabular}{@{}l@{\hskip 0.2in}r@{\hskip 0.2in}r@{\hskip 0.2in}r@{}}
		\toprule
        \toprule
   		Model                     & Weights & Optimized & Fraction \\
		\midrule
		Garment Classifier &  669K & 3M & 1     \\
		DeiT                        & 86M  & 6M & 0.07 \\ 
		Mistral-7b            & 7B     & 12M & 0.002 \\
		\bottomrule
        \bottomrule
	\end{tabular}
	\label{tab:fraction_optimized}
\end{table}

\textbf{Time required} -- The time required for pruning is heavily dependent on these parameters: the block size $n$, the number of epochs \codevar{num\_epochs}, the number of steps per epoch \codevar{num\_steps}, and the size of the data sample used for each step \codevar{batch\_size\_pruning}. In \cref{tab:times} we present example run times. As can be seen in \cref{fig:deit_evolution} and \cref{fig:mistral_7b_evolution} for many of the densities it would have been possible to run significantly fewer epochs and still reach a similar level of accuracy. Nevertheless, we report the total pruning time over all epochs, as was required to obtain the results presented in this work.
\begin{table}[htb]
	\centering
 	\caption{Time required for pruning and evaluation. ``Model'' is the model name, ``Pruning'' is the total time required to prune the model given the parameters chosen for our experiments at density 50\%, ``Evaluation'' is the time required to evaluate the model, and ``Instance'' is the type of Amazon EC2 instance that was used.}
	\begin{tabular}{@{}l@{\hskip 0.1in}l@{\hskip 0.15in}l@{\hskip 0.15in}l@{}}
		\toprule
        \toprule
   		Model              & Pruning & Evaluation & Instance \\
		\midrule
		Garment Classifier &    \phantom{1}2.5 hours  & \phantom{8}0.7 sec & \texttt{g5.4xlarge} \\
		DeiT                        &  17.3 hours & 83.4 sec & \texttt{g5.4xlarge} \\ 
		Mistral-7b          &    \phantom{1}7.9 days   & 53.3 min & \texttt{g5.48xlarge} \\
		\bottomrule
        \bottomrule
	\end{tabular}
	\label{tab:times}
\end{table}
%


\section{Conclusions and Outlook}
\label{sec:conclusions}

In this work we propose a new optimization-based pruning heuristic -- the iterative Combinatorial Brain Surgeon (iCBS). It is based on starting from the result of a one-shot pruning technique, and then iteratively improving on it by solving a series of combinatorial optimization problems over blocks of weights. We have tested this pruner on three models: the Garment Classifier, DeiT, and Mistral-7b. We find that it improves on one-shot pruning techniques substantially, at least for some densities. This work shows that in fact Hessian-based pruning methods can be used for large ML models such as LLMs, but effort (GPUs and time) is required. 

The quality of the results as well as the runtime could improve through hyperparameter tuning, the addition of different weight-scoring methods, or the usage of hardware accelerators (as well as the optimization of our code). The main bottleneck depends on the model, dataset, and parameters, but it is either the calculation of the gradients (which is memory-bound), or the solution of the per-block combinatorial optimization problems (which is compute-bound). For the former, GPU memory has been increasing steadily, and this trend is likely to continue, helping to mitigate this issue. For the latter, use of hardware accelerators may help, e.g., we used a solver that runs on several cores via multi-threading. Utilizing GPUs, FPGAs, or perhaps quantum computers may help to increase the block size $n$ that can be used while returning high-quality results. 

Finally, we highlight possible extensions of this research that extend beyond our present work, such as the following:
\begin{itemize}
\item \textit{Structured pruning} -- It is theoretically possible to apply the formulation used in this work to large elements in a NN, rather than a single weight (a node, a layer, or an attention head). 
\item \textit{Structured sparsity} -- Certain GPUs are able to achieve significant speedup when the sparsity is in a particular pattern, for example 2:4 (two out of four adjacent elements are zero). This consideration could be added to iCBS, for example via an additional set of constraints. 
\item \textit{Additional weight-scoring methods} -- It may be possible to improve either the initial pruning or the subsequent selection by devising better weight-scoring methods. For example, one could use additional information such as the variance or range of the activations and not just the mean \cite{dufort2024maxwell}, or domain-specific knowledge \cite{hossain2024not}. 
\item \textit{Fine-tuning} -- Applying fine-tuning after pruning with iCBS would almost certainly improve the results. Another option is to apply the fine-tuning during the pruning process, perhaps by solving a continuous optimization problem over the quadratic loss approximation as done in Ref.~\cite{yu2022combinatorial}.
\item \textit{Quantum solver} -- The QCBO problem \cref{eq:qcbo_block} could, in principle, be solved by a quantum computer -- at least as a proof of concept today, and on larger scales in the future. This could be done, for example, by reformulation as a QUBO problem and then solving it via the Quantum Approximate Optimization Algorithm (QAOA) \cite{farhi2014quantum}, perhaps using a constraint-preserving mixer \cite{hadfield2017quantum}, or using quantum constrained Hamiltonian optimization techniques (Q-CHOP) \cite{perlin2024q}. 
\item \textit{Pruning quantum circuits} -- Due to limited quantum resources, pruning quantum circuits by removing gates could be beneficial, if this can be done while minimally affecting the measured results~\cite{hu2022quantum}. It may be possible to apply iCBO to this problem, with minimal changes, for example to the pruning of variational quantum circuits~\cite{kulshrestha2024qadaprune}.
\end{itemize}

\begin{acknowledgments}

This work is a collaboration between Fidelity Center for Applied Technology, Fidelity Labs, LLC., and Amazon Quantum Solutions Lab. The authors would like to thank Cece Brooks, Michael Dascal, Cory Thigpen, and Ed Cady  for fruitful discussions. Special thanks to Thomas H\"{a}ner for sharing his implementation of the constrained simulated annealing and constrained branch and bound solvers \cite{haener2024solving}. H.G.K.~would like to thank Shiner Bock for inspiration. The Fidelity publishing approval number for this paper is 1176542.1.0. 

\end{acknowledgments}

\bibliographystyle{unsrtnat}
\bibliography{refs}

\begin{thebibliography}{47}
\providecommand{\natexlab}[1]{#1}
\providecommand{\url}[1]{\texttt{#1}}
\expandafter\ifx\csname urlstyle\endcsname\relax
  \providecommand{\doi}[1]{doi: #1}\else
  \providecommand{\doi}{doi: \begingroup \urlstyle{rm}\Url}\fi

\bibitem[Fedus et~al.(2022)Fedus, Zoph, and Shazeer]{fedus2022switch}
William Fedus, Barret Zoph, and Noam Shazeer.
\newblock Switch transformers: {Scaling} to trillion parameter models with simple and efficient sparsity.
\newblock \emph{Journal of Machine Learning Research}, 23\penalty0 (120):\penalty0 1--39, 2022.

\bibitem[Kaplan et~al.(2020)Kaplan, McCandlish, Henighan, Brown, Chess, Child, Gray, Radford, Wu, and Amodei]{kaplan2020scaling}
Jared Kaplan, Sam McCandlish, Tom Henighan, Tom~B Brown, Benjamin Chess, Rewon Child, Scott Gray, Alec Radford, Jeffrey Wu, and Dario Amodei.
\newblock Scaling laws for neural language models.
\newblock \emph{arXiv preprint arXiv:2001.08361}, 2020.

\bibitem[Cheng et~al.(2024)Cheng, Zhang, and Shi]{cheng2024survey}
Hongrong Cheng, Miao Zhang, and Javen~Qinfeng Shi.
\newblock A survey on deep neural network pruning: Taxonomy, comparison, analysis, and recommendations.
\newblock \emph{IEEE Transactions on Pattern Analysis and Machine Intelligence}, 2024.

\bibitem[Lagunas et~al.(2021)Lagunas, Charlaix, Sanh, and Rush]{lagunas2021block}
Fran{\c{c}}ois Lagunas, Ella Charlaix, Victor Sanh, and Alexander~M Rush.
\newblock Block pruning for faster transformers.
\newblock \emph{arXiv preprint arXiv:2109.04838}, 2021.

\bibitem[Kurtic et~al.(2022)Kurtic, Campos, Nguyen, Frantar, Kurtz, Fineran, Goin, and Alistarh]{kurtic2022optimal}
Eldar Kurtic, Daniel Campos, Tuan Nguyen, Elias Frantar, Mark Kurtz, Benjamin Fineran, Michael Goin, and Dan Alistarh.
\newblock The optimal {BERT} surgeon: {Scalable} and accurate second-order pruning for large language models.
\newblock \emph{arXiv preprint arXiv:2203.07259}, 2022.

\bibitem[Frankle and Carbin(2018)]{frankle2018lottery}
Jonathan Frankle and Michael Carbin.
\newblock The lottery ticket hypothesis: {Finding} sparse, trainable neural networks.
\newblock \emph{arXiv preprint arXiv:1803.03635}, 2018.

\bibitem[Sun et~al.(2023)Sun, Liu, Bair, and Kolter]{sun2023simple}
Mingjie Sun, Zhuang Liu, Anna Bair, and J~Zico Kolter.
\newblock A simple and effective pruning approach for large language models.
\newblock \emph{arXiv preprint arXiv:2306.11695}, 2023.

\bibitem[Frantar and Alistarh(2023)]{frantar2023sparsegpt}
Elias Frantar and Dan Alistarh.
\newblock {SparseGPT}: Massive language models can be accurately pruned in one-shot.
\newblock In \emph{International Conference on Machine Learning}, pages 10323--10337. PMLR, 2023.

\bibitem[Glover et~al.(1993)Glover, Taillard, and Taillard]{glover1993user}
Fred Glover, Eric Taillard, and Eric Taillard.
\newblock A user's guide to tabu search.
\newblock \emph{Annals of operations research}, 41\penalty0 (1):\penalty0 1--28, 1993.

\bibitem[Jiang et~al.(2023)Jiang, Sablayrolles, Mensch, Bamford, Chaplot, Casas, Bressand, Lengyel, Lample, Saulnier, et~al.]{jiang2023mistral}
Albert~Q Jiang, Alexandre Sablayrolles, Arthur Mensch, Chris Bamford, Devendra~Singh Chaplot, Diego de~las Casas, Florian Bressand, Gianna Lengyel, Guillaume Lample, Lucile Saulnier, et~al.
\newblock Mistral 7b.
\newblock \emph{arXiv preprint arXiv:2310.06825}, 2023.

\bibitem[Touvron et~al.(2021)Touvron, Cord, Douze, Massa, Sablayrolles, and J{\'e}gou]{touvron2021training}
Hugo Touvron, Matthieu Cord, Matthijs Douze, Francisco Massa, Alexandre Sablayrolles, and Herv{\'e} J{\'e}gou.
\newblock Training data-efficient image transformers \& distillation through attention.
\newblock In \emph{International conference on machine learning}, pages 10347--10357. PMLR, 2021.

\bibitem[LeCun et~al.(1989)LeCun, Denker, and Solla]{lecun1989optimal}
Yann LeCun, John Denker, and Sara Solla.
\newblock Optimal brain damage.
\newblock \emph{Advances in neural information processing systems}, 2, 1989.

\bibitem[Hassibi et~al.(1993{\natexlab{a}})Hassibi, Stork, and Wolff]{hassibi1993optimal}
Babak Hassibi, David~G Stork, and Gregory~J Wolff.
\newblock Optimal brain surgeon and general network pruning.
\newblock In \emph{IEEE international conference on neural networks}, pages 293--299. IEEE, 1993{\natexlab{a}}.

\bibitem[Yu et~al.(2022)Yu, Serra, Ramalingam, and Zhe]{yu2022combinatorial}
Xin Yu, Thiago Serra, Srikumar Ramalingam, and Shandian Zhe.
\newblock The combinatorial brain surgeon: {Pruning} weights that cancel one another in neural networks.
\newblock In \emph{International Conference on Machine Learning}, pages 25668--25683. PMLR, 2022.

\bibitem[Han et~al.(2015)Han, Pool, Tran, and Dally]{han2015learning}
Song Han, Jeff Pool, John Tran, and William Dally.
\newblock Learning both weights and connections for efficient neural network.
\newblock \emph{Advances in neural information processing systems}, 28, 2015.

\bibitem[Blalock et~al.(2020)Blalock, Gonzalez~Ortiz, Frankle, and Guttag]{blalock2020state}
Davis Blalock, Jose~Javier Gonzalez~Ortiz, Jonathan Frankle, and John Guttag.
\newblock What is the state of neural network pruning?
\newblock \emph{Proceedings of machine learning and systems}, 2:\penalty0 129--146, 2020.

\bibitem[Singh and Alistarh(2020)]{singh2020woodfisher}
Sidak~Pal Singh and Dan Alistarh.
\newblock Woodfisher: Efficient second-order approximation for neural network compression.
\newblock \emph{Advances in Neural Information Processing Systems}, 33:\penalty0 18098--18109, 2020.

\bibitem[Benbaki et~al.(2023)Benbaki, Chen, Meng, Hazimeh, Ponomareva, Zhao, and Mazumder]{benbaki2023fast}
Riade Benbaki, Wenyu Chen, Xiang Meng, Hussein Hazimeh, Natalia Ponomareva, Zhe Zhao, and Rahul Mazumder.
\newblock Fast as {CHITA}: Neural network pruning with combinatorial optimization.
\newblock In \emph{International Conference on Machine Learning}, pages 2031--2049. PMLR, 2023.

\bibitem[Rosenberg et~al.(2016)Rosenberg, Vazifeh, Woods, and Haber]{rosenberg2016building}
Gili Rosenberg, Mohammad Vazifeh, Brad Woods, and Eldad Haber.
\newblock Building an iterative heuristic solver for a quantum annealer.
\newblock \emph{Computational Optimization and Applications}, 65:\penalty0 845--869, 2016.

\bibitem[Zintchenko et~al.(2015)Zintchenko, Hastings, and Troyer]{zintchenko2015local}
Ilia Zintchenko, Matthew~B Hastings, and Matthias Troyer.
\newblock From local to global ground states in {Ising} spin glasses.
\newblock \emph{Physical Review B}, 91\penalty0 (2):\penalty0 024201, 2015.

\bibitem[Booth et~al.(2017)Booth, Reinhardt, and Roy]{booth2017partitioning}
Michael Booth, Steven~P Reinhardt, and Aidan Roy.
\newblock Partitioning optimization problems for hybrid classical.
\newblock \emph{quantum execution. Technical Report}, pages 01--09, 2017.

\bibitem[Dubey et~al.(2024)Dubey, Jauhri, Pandey, Kadian, Al-Dahle, Letman, Mathur, Schelten, Yang, Fan, et~al.]{dubey2024llama}
Abhimanyu Dubey, Abhinav Jauhri, Abhinav Pandey, Abhishek Kadian, Ahmad Al-Dahle, Aiesha Letman, Akhil Mathur, Alan Schelten, Amy Yang, Angela Fan, et~al.
\newblock The {Llama} 3 herd of models.
\newblock \emph{arXiv preprint arXiv:2407.21783}, 2024.

\bibitem[Farhi et~al.(2014)Farhi, Goldstone, and Gutmann]{farhi2014quantum}
Edward Farhi, Jeffrey Goldstone, and Sam Gutmann.
\newblock A quantum approximate optimization algorithm.
\newblock \emph{arXiv preprint arXiv:1411.4028}, 2014.

\bibitem[Hadfield et~al.(2017)Hadfield, Wang, Rieffel, O'Gorman, Venturelli, and Biswas]{hadfield2017quantum}
Stuart Hadfield, Zhihui Wang, Eleanor~G Rieffel, Bryan O'Gorman, Davide Venturelli, and Rupak Biswas.
\newblock Quantum approximate optimization with hard and soft constraints.
\newblock In \emph{Proceedings of the Second International Workshop on Post Moores Era Supercomputing}, pages 15--21, 2017.

\bibitem[Perlin et~al.(2024)Perlin, Shaydulin, Hall, Minssen, Li, Dubey, Rines, Anschuetz, Pistoia, and Gokhale]{perlin2024q}
Michael~A Perlin, Ruslan Shaydulin, Benjamin~P Hall, Pierre Minssen, Changhao Li, Kabir Dubey, Rich Rines, Eric~R Anschuetz, Marco Pistoia, and Pranav Gokhale.
\newblock {Q-CHOP}: {Quantum} constrained hamiltonian optimization.
\newblock \emph{arXiv preprint arXiv:2403.05653}, 2024.

\bibitem[Hassibi et~al.(1993{\natexlab{b}})Hassibi, Stork, and Wolff]{hassibi1993optimalextensions}
Babak Hassibi, David Stork, and Gregory Wolff.
\newblock Optimal brain surgeon: {Extensions} and performance comparisons.
\newblock \emph{Advances in neural information processing systems}, 6, 1993{\natexlab{b}}.

\bibitem[Boettcher(2004)]{boettcher2004extremal}
Stefan Boettcher.
\newblock Extremal optimization.
\newblock \emph{New optimization algorithms in physics}, pages 227--251, 2004.

\bibitem[H{\"a}ner et~al.(2024)H{\"a}ner, Booth, Borujeni, and Zhu]{haener2024solving}
Thomas H{\"a}ner, Kyle E.~C. Booth, Sima~E. Borujeni, and Elton~Yechao Zhu.
\newblock Solving {QUBOs} with a quantum-amenable branch and bound method.
\newblock \emph{arXiv preprint arXiv:2407.20185}, 2024.

\bibitem[Vaswani(2017)]{vaswani2017attention}
A~Vaswani.
\newblock Attention is all you need.
\newblock \emph{Advances in Neural Information Processing Systems}, 2017.

\bibitem[Clark et~al.(2019)Clark, Lee, Chang, Kwiatkowski, Collins, and Toutanova]{clark2019boolq}
Christopher Clark, Kenton Lee, Ming-Wei Chang, Tom Kwiatkowski, Michael Collins, and Kristina Toutanova.
\newblock {BoolQ}: Exploring the surprising difficulty of natural yes/no questions.
\newblock \emph{arXiv preprint arXiv:1905.10044}, 2019.

\bibitem[Wang et~al.(2018)Wang, Singh, Michael, Hill, Levy, and Bowman]{wang2018glue}
Alex Wang, Amanpreet Singh, Julian Michael, Felix Hill, Omer Levy, and Samuel~R Bowman.
\newblock {GLUE}: A multi-task benchmark and analysis platform for natural language understanding.
\newblock \emph{arXiv preprint arXiv:1804.07461}, 2018.

\bibitem[Zellers et~al.(2019)Zellers, Holtzman, Bisk, Farhadi, and Choi]{zellers2019hellaswag}
Rowan Zellers, Ari Holtzman, Yonatan Bisk, Ali Farhadi, and Yejin Choi.
\newblock Hellaswag: Can a machine really finish your sentence?
\newblock \emph{arXiv preprint arXiv:1905.07830}, 2019.

\bibitem[Sakaguchi et~al.(2021)Sakaguchi, Bras, Bhagavatula, and Choi]{sakaguchi2021winogrande}
Keisuke Sakaguchi, Ronan~Le Bras, Chandra Bhagavatula, and Yejin Choi.
\newblock Winogrande: An adversarial winograd schema challenge at scale.
\newblock \emph{Communications of the ACM}, 64\penalty0 (9):\penalty0 99--106, 2021.

\bibitem[Clark et~al.(2018)Clark, Cowhey, Etzioni, Khot, Sabharwal, Schoenick, and Tafjord]{clark2018think}
Peter Clark, Isaac Cowhey, Oren Etzioni, Tushar Khot, Ashish Sabharwal, Carissa Schoenick, and Oyvind Tafjord.
\newblock Think you have solved question answering? try arc, the ai2 reasoning challenge.
\newblock \emph{arXiv preprint arXiv:1803.05457}, 2018.

\bibitem[Mihaylov et~al.(2018)Mihaylov, Clark, Khot, and Sabharwal]{mihaylov2018can}
Todor Mihaylov, Peter Clark, Tushar Khot, and Ashish Sabharwal.
\newblock Can a suit of armor conduct electricity? a new dataset for open book question answering.
\newblock \emph{arXiv preprint arXiv:1809.02789}, 2018.

\bibitem[Xiao et~al.(2017)Xiao, Rasul, and Vollgraf]{xiao2017fashion}
Han Xiao, Kashif Rasul, and Roland Vollgraf.
\newblock {Fashion-MNIST}: a novel image dataset for benchmarking machine learning algorithms.
\newblock \emph{arXiv preprint arXiv:1708.07747}, 2017.

\bibitem[Deng et~al.(2009)Deng, Dong, Socher, Li, Li, and Fei-Fei]{deng2009imagenet}
Jia Deng, Wei Dong, Richard Socher, Li-Jia Li, Kai Li, and Li~Fei-Fei.
\newblock Imagenet: A large-scale hierarchical image database.
\newblock In \emph{2009 IEEE conference on computer vision and pattern recognition}, pages 248--255. Ieee, 2009.

\bibitem[Raffel et~al.(2020)Raffel, Shazeer, Roberts, Lee, Narang, Matena, Zhou, Li, and Liu]{raffel2020exploring}
Colin Raffel, Noam Shazeer, Adam Roberts, Katherine Lee, Sharan Narang, Michael Matena, Yanqi Zhou, Wei Li, and Peter~J Liu.
\newblock Exploring the limits of transfer learning with a unified text-to-text transformer.
\newblock \emph{Journal of machine learning research}, 21\penalty0 (140):\penalty0 1--67, 2020.

\bibitem[Gao et~al.(2023)Gao, Tow, Abbasi, Biderman, Black, DiPofi, Foster, Golding, Hsu, Le~Noac'h, Li, McDonell, Muennighoff, Ociepa, Phang, Reynolds, Schoelkopf, Skowron, Sutawika, Tang, Thite, Wang, Wang, and Zou]{eval-harness}
Leo Gao, Jonathan Tow, Baber Abbasi, Stella Biderman, Sid Black, Anthony DiPofi, Charles Foster, Laurence Golding, Jeffrey Hsu, Alain Le~Noac'h, Haonan Li, Kyle McDonell, Niklas Muennighoff, Chris Ociepa, Jason Phang, Laria Reynolds, Hailey Schoelkopf, Aviya Skowron, Lintang Sutawika, Eric Tang, Anish Thite, Ben Wang, Kevin Wang, and Andy Zou.
\newblock A framework for few-shot language model evaluation, 12 2023.
\newblock URL \url{https://zenodo.org/records/10256836}.

\bibitem[Liaw et~al.(2018)Liaw, Liang, Nishihara, Moritz, Gonzalez, and Stoica]{liaw2018tune}
Richard Liaw, Eric Liang, Robert Nishihara, Philipp Moritz, Joseph~E Gonzalez, and Ion Stoica.
\newblock Tune: A research platform for distributed model selection and training.
\newblock \emph{arXiv preprint arXiv:1807.05118}, 2018.

\bibitem[Bergstra et~al.(2013)Bergstra, Yamins, and Cox]{bergstra2013making}
James Bergstra, Daniel Yamins, and David Cox.
\newblock Making a science of model search: Hyperparameter optimization in hundreds of dimensions for vision architectures.
\newblock In \emph{International conference on machine learning}, pages 115--123. PMLR, 2013.

\bibitem[Paszke et~al.(2019)Paszke, Gross, Massa, Lerer, Bradbury, Chanan, Killeen, Lin, Gimelshein, Antiga, et~al.]{paszke2019pytorch}
Adam Paszke, Sam Gross, Francisco Massa, Adam Lerer, James Bradbury, Gregory Chanan, Trevor Killeen, Zeming Lin, Natalia Gimelshein, Luca Antiga, et~al.
\newblock Pytorch: An imperative style, high-performance deep learning library.
\newblock \emph{Advances in neural information processing systems}, 32, 2019.

\bibitem[Dufort-Labb{\'e} et~al.(2024)Dufort-Labb{\'e}, D'Oro, Nikishin, Pascanu, Bacon, and Baratin]{dufort2024maxwell}
Simon Dufort-Labb{\'e}, Pierluca D'Oro, Evgenii Nikishin, Razvan Pascanu, Pierre-Luc Bacon, and Aristide Baratin.
\newblock Maxwell's demon at work: {Efficient} pruning by leveraging saturation of neurons.
\newblock \emph{arXiv preprint arXiv:2403.07688}, 2024.

\bibitem[Hossain et~al.(2024)Hossain, Fischer, Burkholz, and Quackenbush]{hossain2024not}
Intekhab Hossain, Jonas Fischer, Rebekka Burkholz, and John Quackenbush.
\newblock Not all tickets are equal and we know it: {Guiding} pruning with domain-specific knowledge.
\newblock \emph{arXiv preprint arXiv:2403.04805}, 2024.

\bibitem[Hu et~al.(2022)Hu, Dong, Wang, Lin, Wang, and Jiang]{hu2022quantum}
Zhirui Hu, Peiyan Dong, Zhepeng Wang, Youzuo Lin, Yanzhi Wang, and Weiwen Jiang.
\newblock Quantum neural network compression.
\newblock In \emph{Proceedings of the 41st IEEE/ACM International Conference on Computer-Aided Design}, pages 1--9, 2022.

\bibitem[Kulshrestha et~al.(2024)Kulshrestha, Liu, Ushijima-Mwesigwa, Bach, and Safro]{kulshrestha2024qadaprune}
Ankit Kulshrestha, Xiaoyuan Liu, Hayato Ushijima-Mwesigwa, Bao Bach, and Ilya Safro.
\newblock Qadaprune: Adaptive parameter pruning for training variational quantum circuits.
\newblock \emph{arXiv preprint arXiv:2408.13352}, 2024.

\bibitem[Sarge et~al.(2019)Sarge, Andersch, Fabel, Micikevicius, and Tran]{sarge2019tips}
Valerie Sarge, Michael Andersch, Lynsey Fabel, Paulius Micikevicius, and John Tran.
\newblock Tips for optimizing gpu performance using tensor cores, 2019.

\end{thebibliography}

\appendix

\section{Experimental parameters}
\label{sec:parameters}

The parameter values used for each of the experiments are presented in \cref{tab:parameters}. The parameters are described below, as well as some general guidelines for setting their values. The best guide is often empirical -- via hyperparameter tuning (as we did for the Garment Classifier and DeiT models), but this may be prohibitive for large models/datasets (such as LLMs):
\begin{itemize}
\item \codevar{num\_epochs} -- The number of epochs done by the pruner (the outer loop). The model is evaluated on the validation data at the end of each epoch. This number was selected to be large enough that the evolution of the pruner could be observed, but not so large that the evaluation would slow the experiment down. 
\item \codevar{num\_steps} -- The number of steps done by the pruner in each epoch (the inner loop). Note that each epoch does not necessarily involve all the training data (our experiments involved only a subset, except for the Garment Classifier experiments) -- the number of samples used in each epoch is equal to the product of \codevar{num\_steps} and \codevar{batch\_size\_pruning}. We chose a value large enough that a noticeable change could be seen to the validation accuracy, but not so large such that the time required for a single epoch would be prohibitive for observing the evolution of the results of the pruning. 
\item \codevar{init\_method} -- The weight-scoring method used to do the initial pruning. We found that the best performing one-pass method for each model/dataset (determined experimentally) was also generally the best initial pruning method to use with iCBS.
\item \codevar{selection\_method} -- The weight-scoring method used to select the block of weights to optimize over in each step. Our experiments indicated that the gradient method is the best selection method for the models/datasets we used. 
\item \codevar{block\_size} ($n$) -- The number of weights selected for each block. The larger this value, the more weights can be updated in a single step. This is also the number of variables in the optimization problem that must be solved, so it has a strong bearing on the run time. Selected to be the largest number for which a noticeable improvement to the validation accuracy was observed without adversely affecting the experiment time. 
\item \codevar{num\_restarts} -- The constrained simulated annealing solver performs \codevar{num\_restarts} independent starts in parallel, one on each CPU, to solve the optimization problem that is constructed in each step. Even though the starts are done in parallel, there is still some overhead, so we chose the smallest number for which a noticeable improvement to the validation accuracy was observed. Generally, we expect that as the \codevar{block\_size} is increased, the total effort of the solver must be increased to maintain the same quality. The effort is the product of the number of starts and number of sweeps. The latter was held constant in our experiments, as was the temperature schedule, due to an abundance of tunable parameters, and they are not described here in detail. 
\item \codevar{batch\_size\_evaluation} -- The number of samples in each evaluation batch -- affects the speed and GPU memory required for evaluation, but regardless of the number chosen, the model is evaluated on all validation samples. We chose the largest number that still provided a noticeable speedup and did not result in an out of memory (OOM) error.  
\item \codevar{batch\_size\_pruning} -- The number of samples used to calculate the gradients and Hessian in each step. It needs to be large enough to capture the diversity of the dataset sufficiently, but the larger it is the more time will be required for the gradient calculation. We chose the largest number that still provided a noticeable improvement to the validation accuracy while still not requiring an inordinate amount of time. 
\item \codevar{batch\_size\_calibration} -- The number of samples used for the initial pruning by the Wanda and gradient weight-scoring methods. We chose the largest number that still provided a noticeable improvement to the validation accuracy. For the LLMs we used 128, as done in Ref.~\cite{sun2023simple}.
\item \codevar{max\_batch\_size} -- The maximum sub-batch size used in forward passes -- which are required to collect the activations for Wanda, as well as the maximum sub-batch size used for gradient calculations. Affects the speed and GPU memory required for these operations, but regardless of the number chosen, all respective samples are included. We chose the largest number that provided a noticeable speedup, and did not result in an OOM error. 
\item \codevar{grad\_multiplier} ($\alpha$) -- The coefficient multiplying the gradient term in the per-block optimization problem. Hyperparameter optimization was used to select a value from a continuous range and suggested that the dependence of the results on this value is not strong, as long as it is not too large/small. 
\item \codevar{ridge\_multiplier} ($\lambda$) -- The coefficient multiplying the ridge term in the per-block optimization problem. Hyperparameter optimization was used to select a value from a discrete set of small powers of 10 and suggested that the dependence of the results on this value is not strong, as long as it is not too large/small. 
\item \codevar{tabu\_frac} -- The maximum length of each tabu list is \codevar{tabu\_frac} of the number of weights in that layer. Hyperparameter optimization was used to select a value from a continuous range and suggested that the dependence of the results on this value is not strong, as long as it is not too large/small. 
\item \codevar{fix\_frac\_prune} -- The fraction of weights with the lowest scores in the initially pruned set that are fixed to always be pruned. Hyperparameter optimization was used to select a value from a continuous range and suggested that the dependence of the results on this value is not strong, as long as it is not too large/small. 
\item \codevar{fix\_frac\_keep}  -- The fraction of weights with the highest scores in the initially kept set that are fixed to always be kept. Hyperparameter optimization was used to select a value from a continuous range and suggested that the dependence of the results on this value is not strong, as long as it is not too large/small. 
\end{itemize}

\begin{table*}[htb]
	\centering
 	\caption{The parameters used in our experiments. }
	\begin{tabular}{@{}l@{\hskip 0.2in}r@{\hskip 0.2in}r@{\hskip 0.2in}r@{}}
		\toprule
        \toprule
   		Parameter & Garment Classifier & DeiT & Mistral-7b \\ 
		\midrule
		\codevar{num\_epochs} & 10 & 12 & 10 \\
		\codevar{num\_steps} & 300 & 500 & 300 \\
		\codevar{init\_method} & Magnitude (per layer) & Wanda (per layer) & Wanda (per output) \\
		\codevar{selection\_method} & Gradient (per layer) & Gradient (per layer) & Gradient (per layer) \\
		\codevar{block\_size} ($n$) & 1024 & 1024 & 4096 \\
        \codevar{num\_restarts}     &   10 &   10 &   20 \\
		\codevar{batch\_size\_evaluation} & 4096 & 64 & 1 \\
		\codevar{batch\_size\_pruning} & 2000 & 1024 & 16 \\
		\codevar{batch\_size\_calibration} & 4096 & 4096 & 128 \\
		\codevar{max\_batch\_size} & None & 64 & 1 \\
		\codevar{grad\_multiplier} ($\alpha$) & 0.75 & 0.75 & 0.75 \\
	 	\codevar{ridge\_multiplier} ($\lambda$) & 0.001 & 0.001 & 0.001 \\
		\codevar{tabu\_frac} & 0.40 & 0.40 & 0.40 \\
		\codevar{fix\_frac\_prune} & 0.42 & 0.42 & 0.42 \\
		\codevar{fix\_frac\_keep} & 0.35 & 0.35 & 0.35 \\
		\bottomrule
        \bottomrule
	\end{tabular}
	\label{tab:parameters}
\end{table*}

\section{Practical tricks and tips}

In this section we gather tricks and tips (especially useful for work involving LLMs) that we have come across during our research and have used or at least experimented with. We have found that typical scientific works utilize many of these, but seldom mention them. Therefore, we hope that gathering a selection of them here will prove insightful to the reader, especially those new to this field. This write up is not meant to be comprehensive, and additional resources are provided to the interested reader. Some of the tips may be specific to Python and/or \textsc{PyTorch}, but we expect that most would be relevant to other languages/libraries. 

\begin{itemize}
\item \textbf{Setting the batch size} -- There are algorithm-specific considerations to the sizing of batches. For example, in our case we expect that we need to set the calibration and pruning batch sizes large enough, such that they capture sufficient diversity from the dataset. This needs to be traded off against the additional size generally causing the run time to be longer. In addition to this, there are device-specific considerations -- there is a long standing custom of setting batch sizes to be a power of 2, rooted in various hardware considerations. Recent NVIDIA GPUs appear benefit from setting batch sizes to be multiples of 8 or 16, depending on the precision \cite{sarge2019tips}. 

\item \textbf{Sub-batching / gradient accumulation} -- During forward passes, or when calculating the mean gradient, or per-sample gradient, memory usage can be reduced by sub-batching. By this we mean that an operation to be performed on a batch can be broken down into smaller (and cheaper) operations done on sub-batches of that batch. In some contexts this is referred to as ``gradient accumulation''. 

\item \textbf{The Dataloader class} -- The DataLoader class is used in \textsc{PyTorch} to load datasets. For several of its parameters, changing the default values may benefit your use case, especially when using GPUs. By default, the data is loaded onto the GPU only by a single process (${\codevar{num\_workers}=0}$). For our use case we found that setting the number of workers to at least 4 was beneficial. It can apparently be beneficial for some use cases to allocate the required memory as pinned memory (${\codevar{pin\_memory}=\mbox{True}}$), rather than the default pageable memory, although we found that for our use case it was not. 

When using multiple workers, if the DataLoader object stores any large data structures in memory, those may get copied across the workers, for example if they are standard Python types. In contrast, \textsc{PyTorch} tensors are shared across the workers by default. This is an issue, for example, with the ImageFolder class of the \textsc{datasets} package, when used with the Imagenet-1K dataset we used in our study. This is because it stores a list of strings containing the filenames of the images in the training set (over 1M of them). We avoided this issue by packing the filenames into a \textsc{PyTorch} byte tensor. 

\item \textbf{Precision timing} -- When transferring data using the above non-blocking option, it's necessary to force a synchronization in \textsc{PyTorch} using \codevar{torch.cuda.synchronize()}. 

\item \textbf{Transferring tensors to the GPU} -- When transferring a \textsc{PyTorch} tensor to the GPU using the \codevar{to()} method, it seems to be good practice to set the ${\codevar{non\_blocking=\mbox{True}}}$ parameter. In cases where there is no synchronization point (like a forward pass) this can allow transferring data while doing other operations, leading to a speedup. If there is a synchronization point, this should not have an effect. 

\item \textbf{Memory management on GPUs} -- We found that it's important to consider carefully which datastructures should be stored on the GPU/CPU, and when they should be transferred. In addition on CPUs, Python's garbage collection can typically be relied on -- it is rare that a user would benefit from invoking it explicitly. However, we found that with GPUs it is often better to explicitly delete tensors from the GPU (using the standard \codevar{del} keyword) when they are no longer needed, and then clear the cache explicitly using \codevar{torch.cuda.empty\_cache()}. 

\item \textbf{Multi-GPU usage} -- For large models, forward and backward passes can be memory-bound -- the Hugging Face Model Memory Calculator is a useful resource for estimating memory requirements. In such cases, it can be useful to break down the model such that the layers are spread out across several GPUs (also referred to as ``sharding''). In our code this was accomplished easily by setting ${\codevar{device\_map=\mbox{"auto"}}}$ when calling \codevar{AutoModelForCausalLM.from\_pretrained()} (which balances the memory usage -- there are other options). When performing forward and backward passes, the data is then transferred from GPU to GPU automatically, as needed. 

\item \textbf{Floating point precision} -- In some cases it makes sense to load the model using a lower floating point precision level (say \codevar{float16}) -- as we did in our project for the LLMs. Some operations, like linear layers, are generally faster at lower precision, while other operations are sensitive to reductions in the dynamic range that come with lower precision, like batch and layer normalization. It is possible to benefit from the best of both worlds for models containing such layers (such as DeiT) by using the Automatic Mixed Precision (AMP) package in \textsc{PyTorch}. More explicitly, we did this in our code by loading the model using \codevar{float32} and wrapping the forward passes and loss calculation with the \codevar{torch.amp.autocast} context manager with the parameters ${\codevar{device\_type="\mbox{cuda}"}}$ and ${\codevar{\mbox{dtype}=\mbox{torch.float16}}}$. 

Typically in our LLM experiments we found the GPUs to be memory-bound while CPUs were not due to the larger RAM available. This can be taken advantage of to increase the dynamic range on the CPU. For example, in our case we loaded the LLMs using \codevar{float16}, to minimize the memory usage on the GPUs. Then when estimating the Hessian -- which requires multiplying the per-sample gradients, we did that on the CPU using \codevar{float32}, to avoid under-flowing and over-flowing issues. 

\item \textbf{Choice of data structures} -- Some standard Python data structures, such as sets, can be very inefficient, which is exposed when they contain a large number of elements. For this reason we minimized our usage of sets and utilized the \textsc{bitarray} library for storing the (binary) information on which weights should be pruned. For tabu lists, a Double Ended Queue (such as provided by \codevar{collections.deque}) can be much more efficient than a list ($\mathcal{O}(1)$ versus $\mathcal{O}(n)$ for appending and popping). 

\item \textbf{Key-value caching} -- LLMs sometimes cache key-value data to speed up inference. However, this requires additional GPU memory that may be in short supply. It's sometimes useful to disable this option, such as by setting ${\codevar{model.config.use\_cache=\mbox{False}}}$ (this option is model dependent). 
\end{itemize}

Additional sources of information:
\begin{itemize}
\item The Performance and Scalability page on the Hugging Face website: \url{https://huggingface.co/docs/transformers/en/performance}
\item The Performance Tuning Guide on the \textsc{PyTorch} website: \url{https://pytorch.org/tutorials/recipes/recipes/tuning_guide.html}
\end{itemize} 

\end{document}